\documentclass[10pt,twocolumn,letterpaper]{article}

\usepackage{cvpr}              % To produce the CAMERA-READY version
\usepackage{customization}

\usepackage{graphicx}
\usepackage{amsmath}
\usepackage{amssymb}
\usepackage{booktabs}
\usepackage{algorithm}
\usepackage{algpseudocode}
\usepackage[accsupp]{axessibility}  % Improves PDF readability for those with disabilities.
\usepackage{color, colortbl} % color
\usepackage{epsfig}
\definecolor{Gray}{gray}{0.875}
\definecolor{LightCyan}{rgb}{0.9,1,1}
\usepackage{tabularx, booktabs, xcolor}
\usepackage{array}
\usepackage{makecell}
\usepackage{algorithmicx,algpseudocode}

\newcommand\blfootnote[1]{%
  \begingroup
  \renewcommand\thefootnote{}\footnote{#1}%
  \addtocounter{footnote}{-1}%
  \endgroup
}

\newcolumntype{Y}{>{\centering\arraybackslash}X}

\definecolor{cvprblue}{rgb}{0.21,0.49,0.74}
\usepackage[pagebackref,breaklinks,colorlinks,allcolors=cvprblue]{hyperref}

\newtheorem{thm}{Theorem}
\newtheorem{prop}[thm]{Proposition}
\theoremstyle{definition}
\newtheorem*{defn*}{Definition}
\newtheorem*{ass*}{Assumption}
\newtheorem*{problem*}{Problem}

\usepackage[capitalize]{cleveref}
\crefname{section}{Sec.}{Secs.}
\Crefname{section}{Section}{Sections}
\Crefname{table}{Table}{Tables}
\crefname{table}{Tab.}{Tabs.}

\def\paperID{10496} % *** Enter the Paper ID here
\def\confName{CVPR}
\def\confYear{2025}

\title{TailedCore: Few-Shot Sampling for Unsupervised Long-Tail Noisy Anomaly Detection}

\author{
    Yoon Gyo Jung$^{*1}$\quad Jaewoo Park$^{*\ddagger2}$\quad Jaeho Yoon$^{*3}$\quad Kuan-Chuan Peng$^{4}$
    \\ 
    Wonchul Kim$^{2}$ \quad Andrew Beng Jin Teoh$^{3}$ \quad Octavia Camps$^{\dagger1}$ \vspace{8pt}\\
    $^{1}$Northeastern University\qquad $^{2}$AiV Co. \qquad $^{3}$Yonsei University \vspace{8pt}
    \\
    $^{4}$Mitsubishi Electric Research Laboratories\vspace{8pt}\\
}

\begin{document}
\maketitle
\blfootnote{$^{*}$Equal contribution. $^{\ddagger}$Project lead. $^{\dagger}$Corresponding author.}

\begin{abstract}
We aim to solve unsupervised anomaly detection in a practical challenging environment where the normal dataset is both contaminated with defective regions and its product class distribution is tailed but unknown. We observe that existing models suffer from tail-versus-noise trade-off where if a model is robust against pixel noise, then its performance deteriorates on tail class samples, and vice versa. To mitigate the issue, we handle the tail class and noise samples independently. To this end, we propose \samp, a novel class size predictor that estimates the class cardinality of samples based on a symmetric assumption on the class-wise distribution of embedding similarities. \samp can be utilized to sample the tail class samples exclusively, allowing to handle them separately. Based on these facets, we build a memory-based anomaly detection model \ours, whose memory both well captures tail class information and is noise-robust. We extensively validate the effectiveness of \ours on the unsupervised long-tail noisy anomaly detection setting, and show that \ours outperforms the state-of-the-art in most settings. Code is available in \href{https://github.com/jungyg/TailedCore}{TailedCore}.
\end{abstract}    
\section{Introduction}

%[testing commands: \JP{Jaewoo's comment}; \YJ{Yoon's comment}; \KP{Kuan-Chuan's comment}, \eg, \ie, \etal, \etc, \ours, \ourso.]

In the complex landscape  of anomaly detection, the challenge often lies in navigating practical scenarios that feature a diverse and imbalanced data distribution \cite{johnson2019survey,chen2022imbalance,cao2024survey,diers2023survey}. 
We delve into a challenging yet highly realistic scenario in anomaly detection, where the training dataset, comprising multiple product classes \cite{you2022unified}, is beset by two significant complications: \textit{contamination with noise} \cite{jiang2022softpatch} and \textit{presence of tailed class distributions} \cite{liu2019large}.
In this context, `tail' (few-shot) classes have limited few-shot samples, unlike `head' (many-shot) classes which are data-rich but might contain samples with defects. Crucially, in this unsupervised setting, the class information of these products remains unknown to the model trainer.
This setting mirrors a common predicament in large-scale industrial applications, where companies are required to manage vast production lines encompassing a wide array of product types, each with varying production rates.

\begin{figure}[t]
\centering
\includegraphics[width=.995\linewidth]{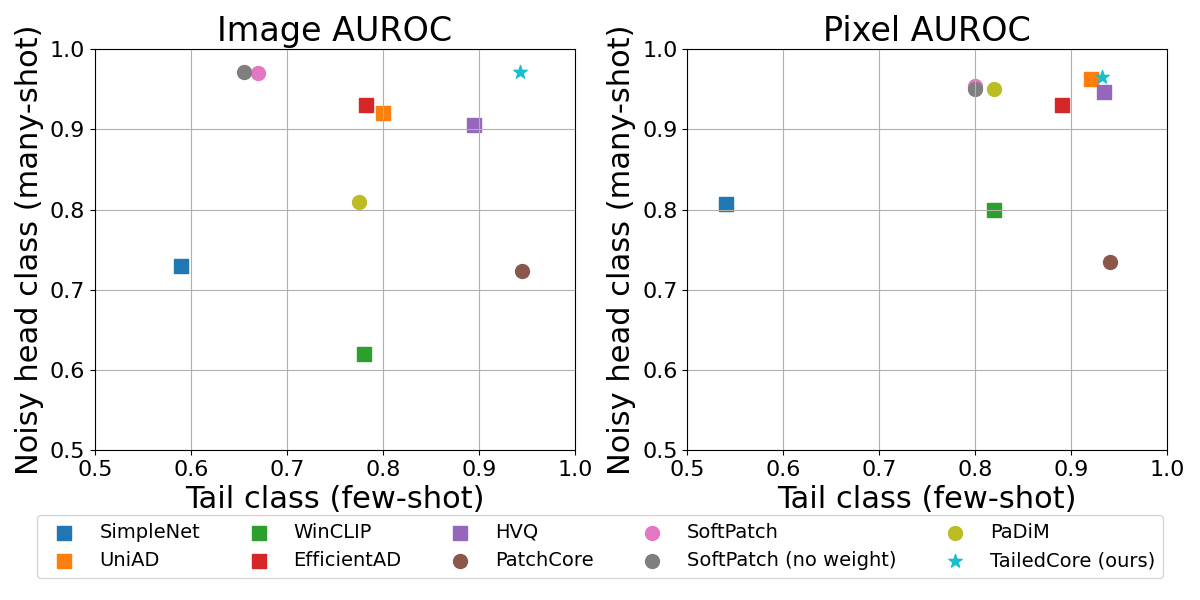}
\caption{
Tail class (x-axis) and noisy head class (y-axis) performance comparison. The tail-versus-noise trade-off is shown across memory-based anomaly detection models (circles), and is more indicative in anomaly classification task evaluated by image-level AUROC (left). 
}
% \KP{In the legend, say \ourso, not just ``ours" for better branding. If possible, please also mark the venue of the SOTA baselines in the legend (see what I did in Tab. 1).} \YJ{Fixed the figure}
\vspace{-.4cm}
\label{fig:bias}
\end{figure}

A critical issue arises in this scenario, which we term the \textit{``tail-versus-noise dilemma.''} The dilemma underscores a fundamental trade-off where \textit{models robust to class imbalance overfit to defect noises}, likely misclassifying many defects as normal. In contrast, existing \textit{models that are robust against defect noise tend to underfit}, missing vital details in tail classes. Fig.~\ref{fig:bias} shows this tail-versus-noise trade-off which is prevalent through existing anomaly detection algorithms. This intrinsic characteristic comes from the fact that few-shot class features are equivalent to the noisy abnormal features in terms of their statistical occurrence. Consequently, existing models find it difficult to strike a balance between avoiding overfitting to noise and effectively capturing the nuances of less represented classes. 

To resolve this challenge, we propose to handle tail classes independently from head classes by exclusively sampling them. To this end, we devise a novel class size predictor \samp that estimates the cardinality of the class of any training sample. For accurate estimation, \samp estimates the class sizes based on a reflective symmetry between inter-class and intra-class similarity distributions in the embedding space. The estimated class sizes of the training samples then enable to sample tail class features exclusively from the training set.

By utilizing the aggregated few-shot class instances, we devise a memory-based anomaly detector \ours, which is built upon the noise-discriminated model by augmenting the tail class patch features sampled with \samp. As the memory of \ours is both clean from noisy defect patches and preserves the few-shot class information, it exhibits robustness against both noise contamination and class imbalance of the training set. 

Our contributions are summarized as follows:
\begin{itemize}
\item We explore a practical and challenging anomaly detection scenario, where a single detection model is trained on multi-class product training samples that involve noise contamination and whose class distribution is long-tailed.

\item We propose a memory-based anomaly detector \ours whose patch memory bank is both noise-free and augmented with the representative patches of few-shot class instances. To build \ours, we introduce \samp that can aggregate few-shot class instances exclusively. \samp is based on a novel class size predictor that estimates the class size of training samples based on the reflective symmetry between inter-class and intra-class class-wise embedding similarity distributions.

\item We conduct extensive analyses and comparative studies on both \ours and few-shot sampler along with the proposed class size predictor on the proposed unsupervised long-tail noisy anomaly detection benchmark. 
    
\end{itemize}

The codes for \ours along with the data generation will be open sourced. We use the term `few-shot' and `tail' interchangeably to denote the class with a small number of samples.
\section{Related works}

\subsection{Anomaly detection in extreme scenarios}
One vitally important aspect of anomaly detection \cite{defard2021padim, roth2022towards} is the environment where these models are tested. This exploration has led to the identification of several challenging scenarios, including multi-class settings with training data spanning multiple product types \cite{you2022unified,lu2023hierarchical}, zero-shot and few-shot settings with minimal training images \cite{jeong2023winclip,zhou2024anomalyclip,fang2023fastrecon,xie2023pushing,li2024musc}, and noisy settings featuring uncleaned training data with defective images \cite{mcintosh2023inter,jiang2022softpatch,chen2022deep}.

\paragraph{Multi-class anomaly detection}
In contrast to single-product settings, anomaly detection models trained on multi-class datasets often encounter ``shortcut learning'' \cite{you2022unified}, leading to low anomaly scores for normal images. This issue is particularly pronounced in reconstruction-based models, which tend to reconstruct all samples, regardless of their normality or abnormality. 
To resolve it, UniAD \cite{you2022unified} employs neighbor masked attention within a transformer model, focusing attention on local parts to prevent shortcut learning.
HVQ \cite{lu2023hierarchical} utilizes vector quantization to discretize the latent space, restricting the network's reconstruction fidelity on normal data distributions.
UniFormaly \cite{lee4586132uniformaly} increases sensitivity to irrelevant patterns by masking less relevant patches and employing top-k matchings with a pretrained foundation model. These multi-class anomaly detection methods do not consider either long-tail classes or contaminated training data setups, and do not work well in the combined ``noisy long-tail'' setup, which is shown in our experimental results.
%\KP{Please add the answers to the following 2 questions here: What's the drawback of the aforesaid methods? What makes \ours better than them?} 

\paragraph{Zero/few-shot anomaly detection}
The inception of few-shot learning benchmarks in anomaly detection marked a significant advancement, with methods leveraging template matching \cite{huang2022registration} and graph representations \cite{xie2023pushing} for minimal training data scenarios.
AnomalyCLIP \cite{zhou2024anomalyclip}, WinCLIP \cite{jeong2023winclip}, and AnomalyGPT \cite{gu2023anomalygpt}, on the other hand, detect anomalies in the few-shot classes based on the acquired knowledge of pretrained foundation models such as CLIP \cite{radford2021learning} and GPT \cite{radford2019language}. However, these methods often require sophisticated prompt engineering or delicate prompt-tuning, and have yet to achieve optimal precision and recall in the single-class training setting with a sufficient amount of training samples. In contrast, our proposed \ours needs none of these process.

\paragraph{Noise anomaly detection}
Dealing with training datasets contaminated with defective samples presents another extreme learning scenario. SoftPatch \cite{jiang2022softpatch} and InReaCh \cite{mcintosh2023inter} focus on purifying memory coresets through outlier detection algorithms and distance association filtering, respectively. IGD \cite{chen2022deep} adopts a robust Gaussian modeling approach, addressing both contaminated training data and small dataset challenges. Yet, these methods do not fully consider scenarios involving both data contamination and class imbalance, unlike our proposed \ours.

\subsection{Long-tail noisy learning}
In the realm of image classification, several studies have focused on learning scenarios involving training sets with label contamination and class imbalance. To address these challenges, the existing literature relies on the robustness and unsupervised aspects of self-supervised learning \cite{zhang2023noisy,lu2023label,fang2022combating,wei2021robust}. The learned embeddings of self-supervised frameworks are less impacted by label contamination and class imbalances. However, these works differ from ours in that they deal with label-level contamination rather than pixel-level defects. Furthermore, these methods require explicit supervised labels to remediate the contamination. In contrast, our work targets noise in pixels without any class label information.

\subsection{Few-shot/outlier sampling}
Both classical clustering methods and outlier detection algorithms can be applied to our task, as they facilitate the identification of few-shot tail class samples, either indirectly or directly.
Unsupervised clustering algorithms \cite{arthur2007k,frey2007clustering,ester1996density} can estimate the cluster sizes but struggle to identify small sample clusters in the imbalanced data. DBSCAN \cite{ester1996density}, on the other hand, captures both clusters and outliers. This however can be problematic when attempting to identify clusters with only one sample (1-shot) as it may be confused between few-shot samples and outliers.
Unlike clustering methods, outlier detection algorithms \cite{angiulli2002fast,breunig2000lof,liu2008isolation,scholkopf1999support} are adept at identifying 1-shot samples which inherently resemble outliers. However, their effectiveness diminishes when a few-shot class contains more than one sample, as these no longer fit the typical outlier profile, potentially leading to incorrect identification of few-shot class samples.
Both algorithm types, due to their inherent design, struggle to accurately predict class/cluster sizes in datasets that are both imbalanced and contaminated. Our proposed class size predictor specifically addresses this issue, aiming to reliably and accurately identify few-shot class samples and enable their exclusive sampling.
Ultimately, the identified few-shot class samples are augmented to the noise-discriminated patch set, constituting the memory-based anomaly detection model \ours, which is robust against both class imbalance and training set noise.

\section{Background}
%We provide a brief background on the memory-based anomaly detection models.

\subsection{Memory-based anomaly detection}
To train a memory-based anomaly detector, a feature extractor $f$ (\eg, WideResNet) extracts a feature map $\phi_i \in \mathbb{R}^{C \times H \times W}$ from each training input image $x_i$. Then, a coreset selection algorithm $\mathcal{S}_{core}$ is applied to the set $P = \{ \phi_i^{(h, \: w)}   \}  $ of patch features $\phi_i^{(h, \: w)}$, resulting in a memory bank 
\begin{equation}
\label{eq:greedy_core}
M_{PatchCore} := \mathcal{S}_{core}(P) \subset P    
\end{equation}
that contains representative patch features of $P$. For instance, PatchCore opts for greedy sampling strategy to obtain the coreset. In the inference stage, the anomaly score $s^{(h, \: w)}$ of the testing image $x$ is computed for each patch feature $\phi^{(h,\: w)}$ by measuring its distance to the memory
\begin{equation}
\label{eq:mem_inf}
score^{(h, \: w)} = dist(\phi^{(h, \: w)}, \ M),
\end{equation}
where $dist$ is the nearest neighbor distance in \cite{roth2022towards}.
The pixel-level anomaly score is computed by up-scaling the score map $score = [score^{(h, \: w)}] \in \mathbb{R}^{H, \: W}$, and its image-level anomaly score is defined as its pixel maximum $\max_{h, \: w} score^{(h, \: w)}$.

\subsection{Noise discrimination of memory}
The training set $X = \{x_i \}$ may be contaminated with defect samples. The naively attained memory bank then contains abnormal patch features, and assigns a low anomaly score on defect images, resulting in poor performance. To resolve this, \cite{jiang2022softpatch} applies a noise discrimination algorithm $\cS_{clean}$ on the patch feature set $P$ prior to coreset sampling:
\begin{equation}
\label{eq:noise_disc}
\cS_{clean}(P) =  \{ \phi_i^{(h, \: w)} : outlier(\phi_i^{(h, \: w)}) < \tau_{0.15} \},
\end{equation}
aggregating only the clean normal patches with low outlier scores. SoftPatch \cite{jiang2022softpatch} exploits local outlier factor (LOF) for outlier scoring, and selects the threshold $\tau_{0.15}$ to remove only $15\%$ of whole patches. In inference, SoftPatch utilizes the memory set $M_{SoftPatch} = \cS_{core} (\cS_{clean} (P))$
\section{Problem and motivation}

\subsection{Problem}
 We consider a practical anomaly detection setting where the normal training data $X =\{x_i\} $ is both multi-class with tail distribution and noise-contaminated. Particularly, each $x_i$ is paired with the product class label $y_i \in \mathcal{Y} = \{1, \: 2, \: \dots\: , \: |\mathcal{Y}| \}$ (\eg, hazelnut, screw, pill, \etc.) and an anomaly label $y^a_i \in \mathcal{Y}^a := \{0, \: 1\}$ with $1$ indicating anomaly. However, this labeling information is \textbf{\textit{unknown}} to the model trainer. The class distribution is long-tailed such that there are tail (\ie, few-shot) classes whose number of training samples is very low; \ie,
\begin{equation}
|C_k| = |\{ x_i : y^c_i = k \}| \le K
\end{equation}
with a very small $K$ (\eg, 1, 4, or 20), where $| \cdot |$ indicates the number of elements in the set. Moreover, some of the training samples in $X$ are anomalies, containing defect blobs in their images. We assume the tail class samples are all normal since in the unsupervised setting its normality is determined by its majority. Overall, the training set $X$ with the above descriptions constitutes \textit{unsupervised tailed noisy anomaly detection} setting. 

\subsection{Motivation: Tail-versus-noise trade-off}

\begin{figure}[t]
\centering
\includegraphics[width=\linewidth]{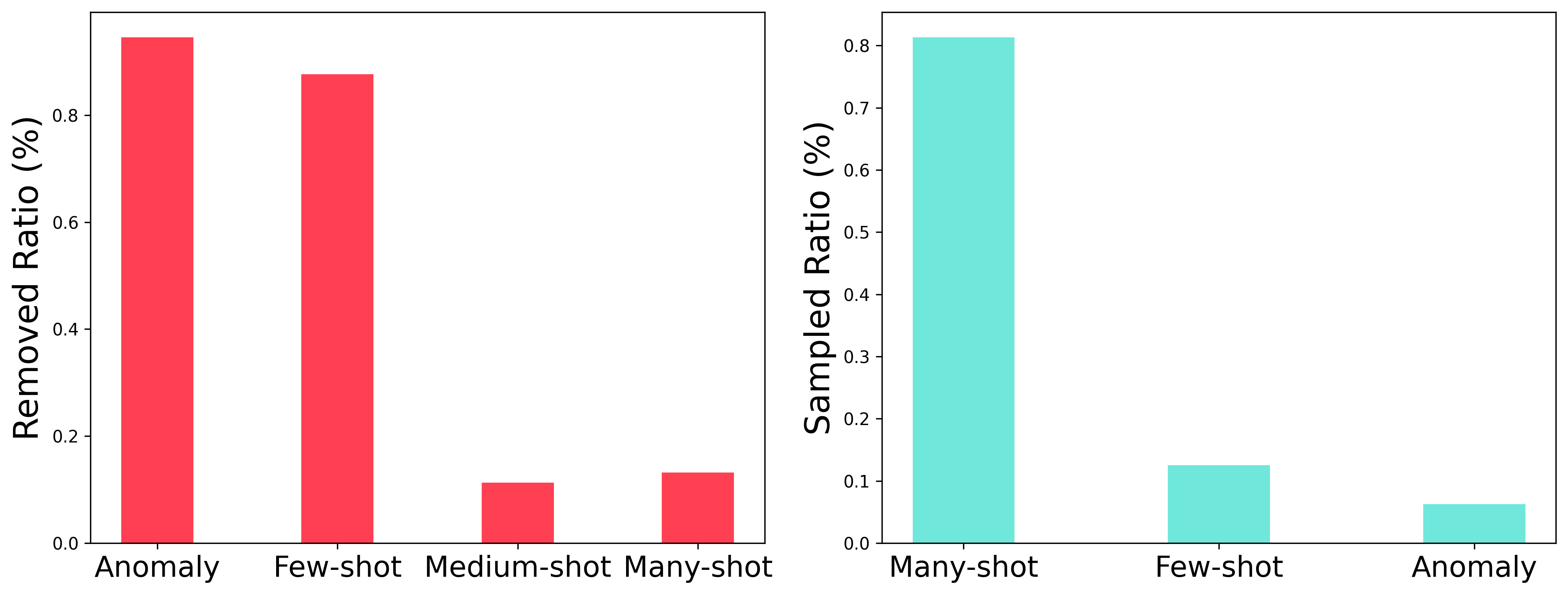}
\caption{
(left)
The ratio of removed patches based on highest outlier scores by Eq.~\eqref{eq:noise_disc}, which shows that most of few-shot class patches are lost.
(right) 
The ratio of sampled patches by greedy coreset sampling from PatchCore which favors both few-shot and anomaly samples.
}
\label{fig:dillema}
\vspace{-.4cm}
\end{figure}

An ideal model to handle a long-tail noisy dataset should selectively sample the few-shot samples while removing noisy samples. However, we observe that the existing models suffer from a trade-off between class-imbalance and noise robustness. For memory-based noise-discriminative models in particular (\eg, SoftPatch), the methods focus on sampling the majority patch features, which often ignores less dominating but important patches. 

As shown in Fig.~\ref{fig:dillema} left, the noise discrimination process removes both defect and few-shot class patches, losing most of the information in the tail distribution. On the other hand, PatchCore samples few-shot samples well due to greedy coreset sampling which captures maximally different features. This sampling principle, however, favors the noise patch features as well as shown in Fig.~\ref{fig:dillema} (right) and the low performance on noisy head classes in Fig.~\ref{fig:bias}. Previous methods, hence, face the tail-versus-noise dilemma described above.

To resolve this, we propose to handle the tail classes independently from the head classes that potentially contain defective local regions. This, however, needs exclusive sampling of tail classes. To achieve this, we use the globally average pooled embedding vector, which is invariant against local pixel variation and rather exclusively corresponds to the object-centric nature of image, \ie, class.

\section{Method}
\label{sec:method}

% \begin{figure}[t]
% \centering
% \includegraphics[width=.75\linewidth]{./figures/method_big.png}
% \caption{
% Model description of (a) PatchCore. (b) SoftPatch. (c) TailedPatch (ours).
% }
% \label{fig:method}
% \end{figure}

% \begin{figure*}[t]
% \centering
% \includegraphics[width=.95\linewidth]{./figures/method2.png}
% \caption{
% Model description of \ourso. \KP{The left figure use the same blue boundary to denote head class normal images and tail class normal images. If possible, I suggest that we use different colors to distinguish them (a total of 3 colors for 3 types of images, and make the color consistent with the dot sample colors on the right and legend colors).}
% }
% \label{fig:method}
% \end{figure*}

\begin{figure*}[tb]
\centering
\begin{subfigure}{0.77\linewidth}
\includegraphics[width=.99\linewidth]{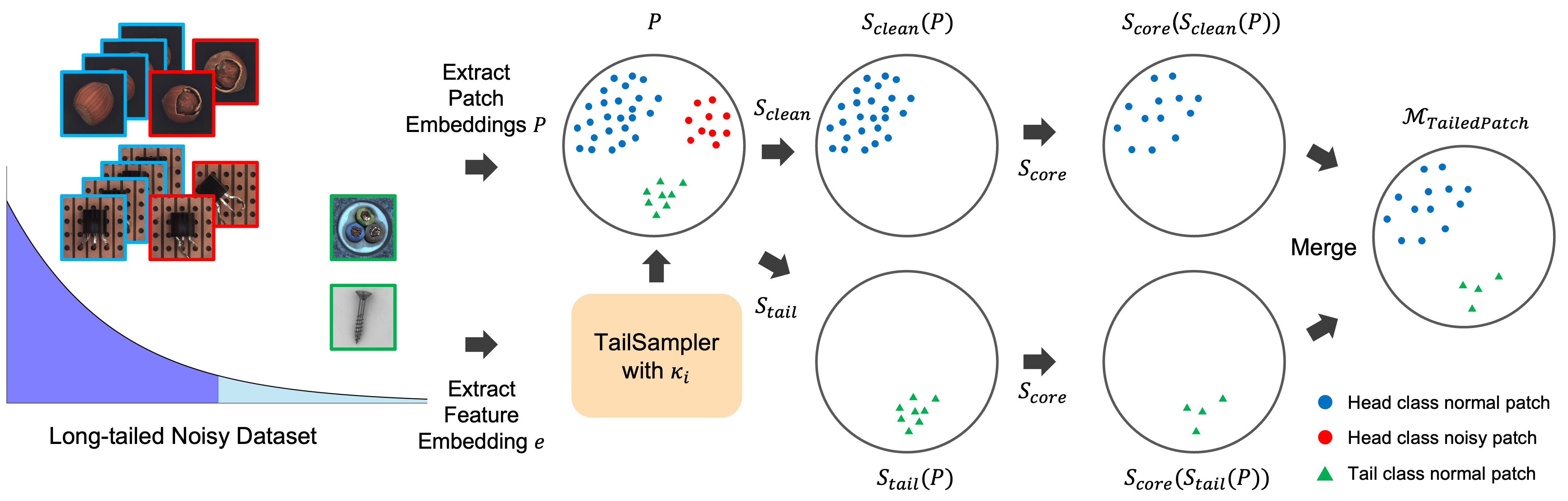}
\caption{}
\end{subfigure}
\hfill
\begin{subfigure}{0.22\linewidth}
\includegraphics[width=.99\linewidth]{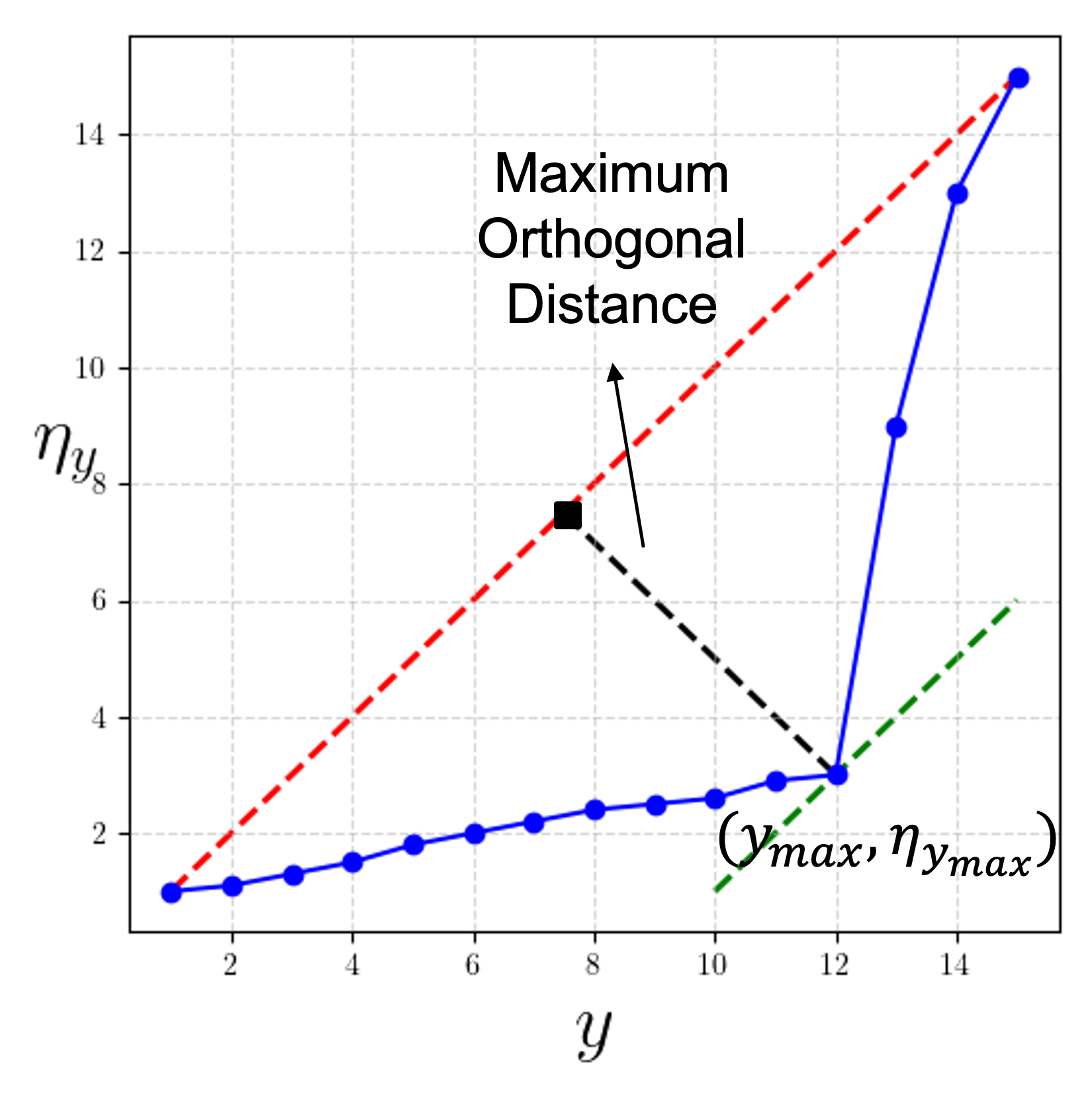}
\caption{}
\end{subfigure}
\caption{
(a) Sampling process description of \ourso and (b) the illustration of how we use the elbow method~\cite{thorndike1953belongs}. 
}
\label{fig:method}
\vspace{-.35cm}
\end{figure*}

%\KP{Fig.~\ref{fig:method} is not referred to anywhere in the main text. If possible, I suggest that we mark the notations (where applicable) we already used to explain \ours in this figure such that it's easier for the readers to follow Sec. 5. On the left of Fig.~\ref{fig:method}, the color boundary of each image is not explained. Maybe we can consider adding some legend nearby. On the right of Fig.~\ref{fig:method}, the point color and shape differences should also be explained with some legend so that the readers don't need to guess what they actually mean.}\YJ{Fixed this part and referred the figure in manuscript.}
% \label{fig:method}
% \end{figure*}

We aim to build a coreset-based anomaly detection model using patch features. We require its memory to aptly capture tail class information and also to be absent of noise patches as much as possible. Particularly, we augment the noise discriminated memory $M_{clean}$ with the tail class memory $M_{tail}$, which contains the patches of few-shot class samples exclusively:
\begin{equation}
M_{TailedCore} = M_{clean} \cup M_{tail}.
\end{equation}
The main challenge lies on sampling the tail class patches exclusively. To this end, we devise a few-shot sampler by estimating the class size of samples based on its embedding (\eg, global average pooling layer of the encoder). Using the estimated class sizes of training samples, the few-shot sampler enables us to exclusively sample tail class samples and the patches thereof.

\subsection{Few-shot sampler: \samp}
\samp first estimates the class size of every training sample. Then, by automatically determining the maximum number of tail classes, denoted by $y_{\max}$, \samp collects the tail class samples' patch features exclusively.

\subsubsection{Class size predictor}

To sample patches exclusively from few-shot class, we identify samples from few-shot classes by estimating their \textit{"class sizes"}, namely the size of class each sample belongs to. To accurately measure the class size, we hypothesize as follows: In the embedding space, samples within an appropriate angle are majorly of the same class, and this angle is likely to be the half of maximum angle to another embedding. In addition, the neighborhoods of neighborhoods within the half angle give accurate class size estimate. Based on these hypotheses, we proceed as follows:

First, we adaptively set angle $\alpha_i$ for each embedding $e_i$ from training samples $x_i$. The coverage of angle $\alpha_i$ decides the neighbors where a large value would assign many neighbors. 
We set the angle to be half of its maximum within the train set, and hypothesize that most of neighbors within the region is likely of the same class. Particularly, let $H_i = \{ e \in Z: \measuredangle(e_i, e) \leq m_i/2\}$ denote the half-angle region where the maximum angle is $m_i := \max_{e \in Z} \measuredangle (e_i, e)$ and the set $Z$ contains all train embeddings. Then, the adaptive angle $\alpha_i$ is defined to contain $p$-th percentile of the half-angle region:
\begin{equation}
\label{eq:radius}
\alpha_i = \measuredangle (e_i, e_{(p \cdot |H_i|)}),
\end{equation}
where the index $(j)$ of $e_{(j)}$ is sorted in the increasing order of the angle $\measuredangle(e_i, e_{(j)})$. We set $p=0.85$ in all experiments unless specified otherwise.

After setting the adaptive angle $\alpha_i$ for each train embedding $e_i$, we estimate its class size based on neighborhoods of neighborhoods. Let $N_{\alpha}(e_i) = \{ e \in Z : \measuredangle(e_i, e) \leq \alpha \} $ denote the neighborhood of $e_i$, which is the set of all train embedding $e$ within the angle $\alpha$ of $e_i$. Then, the class size is estimated by the mode of the sizes of neighborhoods of the neighborhood:
\begin{equation}
\label{eq:cize}
\kappa_i
=
\underset{e \in N_{\alpha_i}(e_i)}{\mathrm{mode}}
(|N_{\alpha(e)}(e)|),
\end{equation}
where $\alpha(e)$ is the adaptive angle with respect to the embedding $e$ belonging to the neighborhood $N_{\alpha_i}(e_i)$ of embedding $e_i$. Using the neighborhoods of neighborhood and mode on it gives more robust estimation of the class size than mere count of the direct neighborhood of $e_i$.

The proposition in Supp.~\ref{supp_sec:theory} shows that the angle of neighborhood given by Eq.~\eqref{eq:radius} corresponds to the decision boundary that maximizes the inter-class separation under regularities.

\subsubsection{Estimation of maximum size of tail class}
After estimating the class size of each training sample by $\kappa_i$, we determine the maximum size for the tail classes based on the elbow technique \cite{thorndike1953belongs}. 
In particular, by utilizing the sample-wise class sizes $\kappa_i$, we first obtain rough estimation of the size of each class set $\eta_y \approx |C_y|$. Then, we estimate the maximum size of tail classes $K_{max}$ by finding a point where the class size most abruptly changes. This optimal shift point is found by the elbow technique.
Concretely, we first estimate the size of each class $y$, $\eta_y \approx |C_y|$, inductively by
\begin{equation}
\begin{split}
\eta_1 &= round \left(
\frac{1}{ \kappa_{(1)} } \sum_{i=1}^{ \kappa_{(1)} }
\kappa_{(i)}
\right),
\quad  \\
\eta_{y+1} &= 
round \left(
\frac{1}{ \kappa_{(\eta_{y} + 1)} }
\sum_{i = \eta_{y} + 1}^{\min ( \kappa_{(\eta_{y} + 1)}, |X|)}
\kappa_{(i)}
\right),
\end{split}
\end{equation}
where the index $(i)$ of $\kappa_{(i)}$ is sorted in the increasing order $\kappa_{(1)} \leq \cdots \leq \kappa_{(|X|)}$.

After we acquire the set $\{(y, \:  \eta_y): y\in(1, \ \dots \ , |\mathcal{Y}|)\}$ through the inductive process above, we define the maximum size of tail classes as the elbow point of the estimated sizes of classes $\eta_y$ since at this point the class size increases abruptly. The elbow is computed by linearly connecting $(1, \: \eta_1)$  and $(|\mathcal{Y}|, \: \eta_{|\mathcal{Y}|})$, and finding the maximum orthogonal distance onto the line from the set of points $\{(y,\eta_y): y\in(1, \ \dots \, |\mathcal{Y}|)\}$ as shown below:
\begin{equation}
K_{\max} := \eta_{y_{\max}} =: elbow(\eta_{1}, \dots, \eta_{|\mathcal{Y}|})
\end{equation}
where
\begin{equation}
\label{eq:elbow}
y_{\max} = 
\argmax_{y \in \{1, \dots, |\mathcal{Y}| \}} \frac{|ay - \eta_y + b|}{\sqrt{a^2+1}}
\end{equation}
and
% \begin{equation}
% \begin{split}
% K_{\max} = \eta_{y_\max} &= elbow(\eta_{1}, \ \dots \ , \ \eta_{|\mathcal{Y}|}), \\
% &:= \argmax_{\eta_y \in \{\eta_1, \dots, \eta_{|\mathcal{Y}|} \}} \frac{|my - \eta_y + b|}{\sqrt{m^2+1}},
% \end{split}
% \label{eq:elbow}
% \end{equation}
where the line $\eta_y = ay +b $ intersects $(1, \: \eta_1)$ and $(|\mathcal{Y}|, \: \eta_{|\mathcal{Y}|})$ with $a=(\eta_{|\mathcal{Y}|} - \eta_1)/(|\mathcal{Y}| - 1)$ and $b = \eta_1 - m$. Fig.~\ref{fig:method} (b) illustrates Eq.~\eqref{eq:elbow}. We choose the elbow point $(y_{max}, \eta_{y_{max}})$ as the maximum size of tail classes $K_{max}$.

% The elbow point finds the point where the class size increases abruptly by finding the maximally long orthogonal distance onto the line connected by $(1, \: \eta_1), (K, \: \eta_K)$ from the set of points $\{(k,\eta_k):k\in(1, \ \dots \, K)\}$.

The maximum size of tail classes $K_{max}$ allows us to sample the patch features of tail classes exclusively:
% \begin{equation}
% \label{eq:tail_sample}
% \mathcal{S}_{tail}(P) = \{ \phi_{i}^{(h,w)} : \mathcal{C}(x_i) \leq K_{max}, \quad x_i \in X  \}.
% \end{equation}
\begin{equation}
\label{eq:tail_sample}
\cS_{tail}(P) = \{ \phi_{i}^{(h,w)} : \kappa_i \leq K_{max} \}.
\end{equation}
with $\kappa_i$ given by Eq.~\eqref{eq:cize}.
%\KP{$cize$ or $cize_r$? Can you make ALL the notations consistent with Eq. 6?}

%%% REVISED (end)

\subsection{Proposed method: \ours}
Our anomaly detection model is built upon the memory bank constructed as below. First, we sample the base memory bank as in \cite{jiang2022softpatch} with noise-discrimination described in Eq.~\eqref{eq:noise_disc}. Then, the base memory is \textit{augmented with \samp} $\mathcal{S}_{tail}$ (Eq.~\eqref{eq:tail_sample}). Overall, \ours's memory bank is defined by $M_{TailedCore} = \cS_{core} ( \cS_{clean} (P)) \cup \cS_{core} ( \cS_{tail} (P) )$,
% % old
%\begin{equation}
%M_{TailedPatch} = \cS_{core} ( \cS_{clean} (P)) \cup \cS_{core} ( \cS_{tail} (P) ),
%\end{equation}
% new
% \begin{equation}
% M_{TailedPatch} = \mathcal{S}_{core} ( \mathcal{S}_{clean} (P \cup \mathcal{S}_{tail} (P)))
% \end{equation}
where $\mathcal{S}_{core}$ is the greedy coreset sampling in Eq.~\eqref{eq:greedy_core}, $\mathcal{S}_{clean}$ is the noise discrimination process that filters out defect patches based on their outlier scores (Eq.~\eqref{eq:noise_disc}), and the few-shot sampler $\mathcal{S}_{tail}$ augments the tail class patches exclusively on the noise-discriminated base memory bank. The whole process is shown in Fig~\ref{fig:method}.
As \ours's patch memory bank is both noise-clean and contains the representative features of few-shot classes, it is robust against both class imbalance and normal sample contamination.
In inference, the anomaly score is obtained by Eq.~\eqref{eq:mem_inf} as in the conventional memory-based detector.

\section{Experiments}
\label{sec:exp}
Section~\ref{sec:exp} consists of dataset description, comparison to the SOTA models, and the detailed ablation study and analysis of \ours's components.

%KP modified table
\begin{table}[t]
\centering
\resizebox{\linewidth}{!}{
\begin{tabular}{r@{\hspace{0.6em}}c@{\hspace{0.6em}}c@{\hspace{0.6em}}cc@{\hspace{0.6em}}c@{\hspace{0.6em}}cc@{\hspace{0.6em}}c@{\hspace{0.6em}}c}
\toprule
tail type &\multicolumn{3}{c}{Pareto} &\multicolumn{3}{c}{step ($K=4$)} &\multicolumn{3}{c}{step ($K=1$)}\\
\cmidrule(r){2-4} \cmidrule(r){5-7} \cmidrule(r){8-10}
class type &$C_t$ &$C_h$ &all &$C_t$ &$C_h$ &all &$C_t$ &$C_h$ &all\\
\midrule
PaDiM~\cite{defard2021padim}~$_{\text{ICPR'21}}$ &82.45 &80.95 &82.06 &77.47 &81.28 &79.19 &71.54 &81.75 &75.63\\
HVQ~\cite{lu2023hierarchical}~$_{\text{NeurIPS'23}}$ &83.46 &80.23 &82.99 &82.01 &85.50 &83.56 &74.15 &90.15 &80.55\\
WinCLIP~\cite{jeong2023winclip}~$_{\text{CVPR'23}}$ &89.35 &90.11 &90.37 &91.60 &88.21 &90.37 &\ul{91.80} &88.23 &90.37\\
AnomalyCLIP~\cite{zhou2024anomalyclip}~$_{\text{ICLR'24}}$ &90.93 &\ul{90.98} &\ul{91.48} &91.82 &90.83 &\ul{91.48} &91.21 &91.90 &\ul{91.48}\\
PatchCore~\cite{roth2022towards}~$_{\text{CVPR'22}}$ &\ul{93.33} &87.59 &89.18 &\ul{92.19} &71.18 &83.83 &86.36 &70.48 &80.01\\
SoftPatch~\cite{jiang2022softpatch}~$_{\text{NeurIPS'22}}$ &84.68 &86.95 &87.71 &67.65 &\bf 97.54 &79.64 &60.66 &\bf 97.49 &75.40\\
\rowcolor{OursColor}\ourso &\bf 96.55 & \bf 95.24 &\bf 96.12 &\bf 95.82 &\ul{95.34} &\bf 95.71 &\bf 93.54 &\ul{95.77} &\bf 94.43\\
\bottomrule
\end{tabular}
}
\caption{Anomaly classification on MVTecAD with image-level AUROC (\%). We report the mean over 5 random seeds for each measurement. Notations: $C_h$ / $C_t$: head / tail classes.  
}
\label{table:sota_mvtec_cls}
\vspace{-.25cm}
\end{table}

%KP modified table
\begin{table}[t]
\centering
\resizebox{\linewidth}{!}{
\begin{tabular}{r@{\hspace{0.6em}}c@{\hspace{0.6em}}c@{\hspace{0.6em}}cc@{\hspace{0.6em}}c@{\hspace{0.6em}}cc@{\hspace{0.6em}}c@{\hspace{0.6em}}c}
\toprule
tail type &\multicolumn{3}{c}{Pareto} &\multicolumn{3}{c}{step ($K=4$)} &\multicolumn{3}{c}{step ($K=1$)}\\
\cmidrule(r){2-4} \cmidrule(r){5-7} \cmidrule(r){8-10}
class type &$C_t$ &$C_h$ &all &$C_t$ &$C_h$ &all &$C_t$ &$C_h$ &all\\
\midrule
PaDiM~\cite{defard2021padim}~$_{\text{ICPR'21}}$ &70.70 &83.35 &78.64 &60.65 &88.93 &72.43 &55.98 &86.75 &68.80\\
HVQ~\cite{lu2023hierarchical}~$_{\text{NeurIPS'23}}$ &73.47 &84.03 &68.25 &68.25 &89.30 &77.02 &61.57 &80.40 &69.42\\
WinCLIP~\cite{jeong2023winclip}~$_{\text{CVPR'23}}$ &73.25 &76.92 &75.47 &75.98 &74.76 &75.47 &78.80 &70.80 &75.47\\
AnomalyCLIP~\cite{zhou2024anomalyclip}~$_{\text{ICLR'24}}$ &81.96 &82.48 &82.05 &82.28 &81.74 &\ul{82.05} &\bf 83.26 &80.34 &\ul{82.05}\\
PatchCore~\cite{roth2022towards}~$_{\text{CVPR'22}}$ &\ul{86.11} &85.73 &85.59 &\ul{83.53} &67.51 &76.85 &79.33 &68.56 &74.84\\
SoftPatch~\cite{jiang2022softpatch}~$_{\text{NeurIPS'22}}$ &78.04 &\ul{92.16} &\ul{86.56} &59.70 &\bf 95.97 &74.81 &52.61 &\bf 94.17 &69.92\\
\rowcolor{OursColor}\ourso &\bf 87.55 & \bf 93.06 &\bf 90.85 &\bf 85.16 &\ul{95.91} &\bf 89.64 &\ul{82.97} &\ul{94.11} &\bf 87.61\\
\bottomrule
\end{tabular}
}
\caption{Anomaly classification on VisA with image-level AUROC (\%). The format and evaluation protocol are the same as Tab.~\ref{table:sota_mvtec_cls}.  
}
\label{table:sota_visa_cls}
\vspace{-.25cm}
\end{table}

%KP modified table
\begin{table}[t]
\centering
\resizebox{\linewidth}{!}{
\begin{tabular}{r@{\hspace{0.6em}}c@{\hspace{0.6em}}c@{\hspace{0.6em}}cc@{\hspace{0.6em}}c@{\hspace{0.6em}}cc@{\hspace{0.6em}}c@{\hspace{0.6em}}c}
\toprule
tail type &\multicolumn{3}{c}{Pareto} &\multicolumn{3}{c}{step ($K=4$)} &\multicolumn{3}{c}{step ($K=1$)}\\
\cmidrule(r){2-4} \cmidrule(r){5-7} \cmidrule(r){8-10}
class type &$C_t$ &$C_h$ &all &$C_t$ &$C_h$ &all &$C_t$ &$C_h$ &all\\
\midrule
PaDiM~\cite{defard2021padim}~$_{\text{ICPR'21}}$ &90.11 &92.66 &91.43 &82.53 &\ul{95.29} &87.67 &78.80 &\ul{95.54} &85.50\\
HVQ~\cite{lu2023hierarchical}~$_{\text{NeurIPS'23}}$ &\ul{93.63} &86.85 &90.55 &90.73 &92.58 &\ul{91.53} &86.36 &95.20 &89.90\\
WinCLIP~\cite{jeong2023winclip}~$_{\text{CVPR'23}}$ &82.03 &84.06 &82.29 &80.60 &84.63 &82.29 &80.16 &85.48 &82.29\\
AnomalyCLIP~\cite{zhou2024anomalyclip}~$_{\text{ICLR'24}}$ &91.24 &91.69 &91.08 &89.96 &92.66 &91.08 &89.34 &93.68 &\ul{91.08}\\
PatchCore~\cite{roth2022towards}~$_{\text{CVPR'22}}$ &93.56 &87.98 &89.93 &\ul{93.54} &72.09 &85.19 &\ul{92.02} &71.35 &83.75\\
SoftPatch~\cite{jiang2022softpatch}~$_{\text{NeurIPS'22}}$ &92.19 &\ul{93.83} &\ul{93.41} &80.98 &\bf 96.49 &87.24 &70.34 &\bf 96.89 &80.99\\
\rowcolor{OursColor}\ourso &\bf 96.08 & \bf 95.01 &\bf 95.29 &\bf 95.56 &93.20 &\bf 94.74 &\bf 94.19 &93.70 &\bf 93.99\\
\bottomrule
\end{tabular}
}
\caption{Anomaly segmentation on MVTecAD with pixel-level AUROC (\%). We report the mean over 5 random seeds for each measurement. Notations: $C_h$ / $C_t$: head / tail classes. 
}
\label{table:sota_mvtec_seg}
\vspace{-.25cm}
\end{table}

%KP modified table
\begin{table}[t]
\centering
\resizebox{\linewidth}{!}{
\begin{tabular}{r@{\hspace{0.6em}}c@{\hspace{0.6em}}c@{\hspace{0.6em}}cc@{\hspace{0.6em}}c@{\hspace{0.6em}}cc@{\hspace{0.6em}}c@{\hspace{0.6em}}c}
\toprule
tail type &\multicolumn{3}{c}{Pareto} &\multicolumn{3}{c}{step ($K=4$)} &\multicolumn{3}{c}{step ($K=1$)}\\
\cmidrule(r){2-4} \cmidrule(r){5-7} \cmidrule(r){8-10}
class type &$C_t$ &$C_h$ &all &$C_t$ &$C_h$ &all &$C_t$ &$C_h$ &all\\
\midrule
PaDiM~\cite{defard2021padim}~$_{\text{ICPR'21}}$ &89.02 &95.10 &82.81 &83.90 &\ul{97.36} &89.51 &82.57 &96.57 &88.40\\
HVQ~\cite{lu2023hierarchical}~$_{\text{NeurIPS'23}}$ &95.27 &\bf 97.60 &\ul{96.71} &93.88 &\bf 98.34 &\ul{95.74} &90.58 &95.51 &92.63\\
WinCLIP~\cite{jeong2023winclip}~$_{\text{CVPR'23}}$ &71.94 &73.97 &73.19 &74.60 &71.21 &73.19 &73.81 &72.32 &73.19\\
AnomalyCLIP~\cite{zhou2024anomalyclip}~$_{\text{ICLR'24}}$ &95.60 &95.46 &95.51 &\ul{95.54} &95.48 &95.51 &\bf 96.16 &94.60 &\ul{95.51}\\
PatchCore~\cite{roth2022towards}~$_{\text{CVPR'22}}$ &\ul{96.84} &87.99 &91.13 &95.39 &62.96 &81.88 &94.11 &65.30 &82.10\\
SoftPatch~\cite{jiang2022softpatch}~$_{\text{NeurIPS'22}}$ &93.20 &96.74 &95.27 &83.95 &97.10 &89.43 &80.73 &\ul{96.82} &87.43\\
\rowcolor{OursColor}\ourso &\bf 97.98 & \ul{97.25} &\bf 97.48 &\bf 96.80 &97.02 &\bf 96.89 &\ul{96.12} &\bf 97.39 &\bf 96.65\\
\bottomrule
\end{tabular}
}
\caption{Anomaly segmentation on VisA with pixel-level AUROC (\%). The format and evaluation protocol are the same as Tab.~\ref{table:sota_mvtec_seg}.
}
\label{table:sota_visa_seg}
\vspace{-.4cm}
\end{table}

\subsection{Datasets for unsupervised long-tail noisy anomaly detection}
We devise the long-tail noisy anomaly detection datasets by modifying the class-wise sample distributions of the widely used anomaly detection datasets MVTecAD \cite{bergmann2019mvtec} and VisA \cite{zou2022spot}. We consider three different types of long-tail distributions. Particularly, datasets with step-shaped tail distribution are devised by making the particular classes to have only a small number of samples; namely $K=1$, and $K=4$, imposing extreme class imblance on the datasets. We name these two datasets by `step $(K=1)$' and `step $(K=4)$'. On the other hand, another realistic type of tail-distribution is considered by making the class distribution to follow the Pareto distribution with the shape parameter $0.6$. For the step-tailed distribution datasets, we impose 40\% of classes to be head classes, containing the original number of training samples. Thus, in MVTecAD, 6 of 15 classes are head classes. For the Pareto tailed distribution, we regard as few-shot (\ie, tail classes) the classes whose number of samples is less than 20 \cite{liu2019large}.
To impose noisy condition on the datasets, the anomaly samples (\ie, the samples with defective regions) are added to the training samples such that the noisy samples constitute 10\% of the training set following the overlap protocol in SoftPatch~\cite{jiang2022softpatch} for fair comparison. 
In all cases, the testing sets of datasets remain the same as the original MVTecAD and VisA. A more detailed configuration is given in Supp.~\ref{supp_sec:dataset}.
%\KP{Where do those abnormal training examples come from for the MVTecAD dataset? Are they from the MVTecAD testing set because its training set only has normal examples? If so, then part of the training set (abnormal training images) overlaps with the testing set, which is an issue.}
%\YJ{fixed this part. should i explain more that noise comes from test set?} \KP{No need to. I've already added that this is for fair comparison w/ SoftPatch.}

\subsection{Comparison to the state-of-the-art (SOTA)}

We compare \ours to the SOTA anomaly detection methods in the unsupervised long-tail noisy settings. We carefully select our baselines. HVQ excels in multi-class learning. PaDiM and SoftPatch are robust against noises as PaDiM is based on Gaussian modeling of the data, while SoftPatch explicitly removes the defect patches by outlier scoring. Using foundation models, WinClip and AnoamlyClip are strong for zero-shot learning. PatchCore are near-perfect in the single-class anomaly detection settings. We show the full experimental results in Supp.~\ref{supp_sec:exp}.
All models are trained in the multi-class setting; namely, one model for one dataset with all classes rather than one model per class. The detection performance, however, is measured for each product class under image-level and pixel-level AUROC, and averaged for reporting and comparison. For comparison, we consider models that are specifically robust against noise, strong on multi-class training, and foundation models that can exhibit zero/few-shot learning capability.

Tabs.~\ref{table:sota_mvtec_cls}, \ref{table:sota_visa_cls}, \ref{table:sota_mvtec_seg}, and \ref{table:sota_visa_seg} show the superiority of \ours across all types of long-tail distributions. As anticipated, all baseline models encounter the tail-versus-noise trade-off, resulting in suboptimal performance either on tail classes with limited samples or on head classes with noisy training samples. These trends become more pronounced in highly unbalanced datasets, where the few-shot classes contain only 1 or 4 samples.

\subsection{Ablation study and analysis}

\subsubsection{Tail (few-shot) sampling}
We evaluate how accurately our proposed algorithm $\mathcal{S}_{tail}$ can sample tail (few-shot) class samples exclusively without including noise samples. First of all, we measure the ratio of mis-included tail class samples \samp which is the ratio of not sampled tail class samples to total tail class samples. A single mis-inclusion can vitally damage anomaly detection performance especially in case of the one-shot setting $K=1$ as the model wouldn't be able to utilize the only sample of the class.

Additionally, the exclusive aspect of few-shot sampling is crucial; the few-shot sampled set shouldn't include anomaly and head class samples as possible. Included anomaly samples would harm the head class performance and head class sampled by few-shot sampler might cause class imbalance issue as well as the possibility of sampling anomaly from head classes. Hence, we measure the ratios of included anomalies and head-class samples along with AUROC which is measured on the outlier scores derived by a given algorithm and ground truth binary labels that indicate whether a given sample is tail class or head class. For \samp, the outlier score of each train sample is given as the negative of predicted class size. The included anomaly ratio is defined as the ratio of number of anomaly samples to the number of whole anomaly samples. The ratio of included head-class sample is defined as the number of included head class to the number of included tail class samples.
%\KP{For branding purposes, I suggest that you give a name for $\mathcal{S}_{tail}$ if you'd like to sell it as a sampling algorithm such that you can claim that we propose a novel sampling method [method\_name].}
%\KP{Is the previous statement supported by any of your exp. results? If so, please guide the readers to that evidence before making that statement.}
%\YJ{please take a look at the previous statement whether it sounds reasonable}
% The described metrics are detailed in Supp.~\ref{supp_sec:analysis}. 
%\KP{Why the numbers of those samples should be as small as possible? This is not obvious for the readers. Please explain. What's the drawback if you include many of those examples?} \YJ{}

For extensive comparison, we compare \samp with widely used clustering algorithms \cite{frey2007clustering,arthur2007k,ester1996density}, outlier detectors \cite{scholkopf1999support,liu2008isolation,breunig2000lof,angiulli2002fast}, and density estimators \cite{dempster1977maximum,parzen1962estimation}. These methods provide clustering size and/or outlier score/prediction. In cases when only outlier scores are available, we apply the elbow method to induce deterministic prediction of whether a given sample is few-shot or not. We use default settings provided by Sklearn \cite{scikit-learn} and PyOD \cite{zhao2019pyod}. For all methods, we utilize the embedding vectors of WideResNet-50 \cite{zagoruyko2016wide}. Experiments with different architectures and details are provided in Supp.~\ref{supp_sec:analysis}. 

The combined ratio, which is the sum of missing tail samples and included anomaly samples, indicates that \samp is most effective for exclusively sampling few-shot class samples. The most critical factor is to not miss few-shot class samples. On this ratio, \samp achieves the lowest rate except for the algorithms that trivially sample all few-shot class samples. We find that most of the classical algorithms for clustering and outlier detection are not usable as they either miss too many few-shot samples or includes a large number of anomaly samples. We note that the class size predictor in our few-shot samples resembles DBSCAN and Affinity Propagation, respectively, in terms of using neighborhoods and voting mechanism. None of these methods however have instance-specific parameters that are adaptively adjusted across different samples.

\begin{figure}[tb]
\centering
\begin{subfigure}{0.49\linewidth}
\includegraphics[width=.99\linewidth]{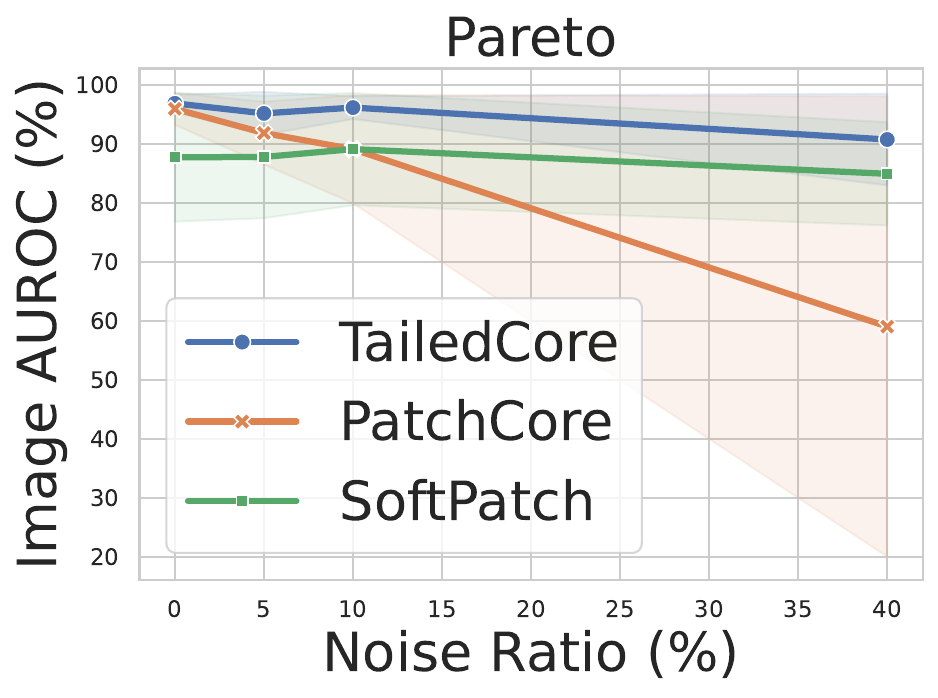}
\caption{}
\end{subfigure}
\hfill
\begin{subfigure}{0.49\linewidth}
\includegraphics[width=.99\linewidth]{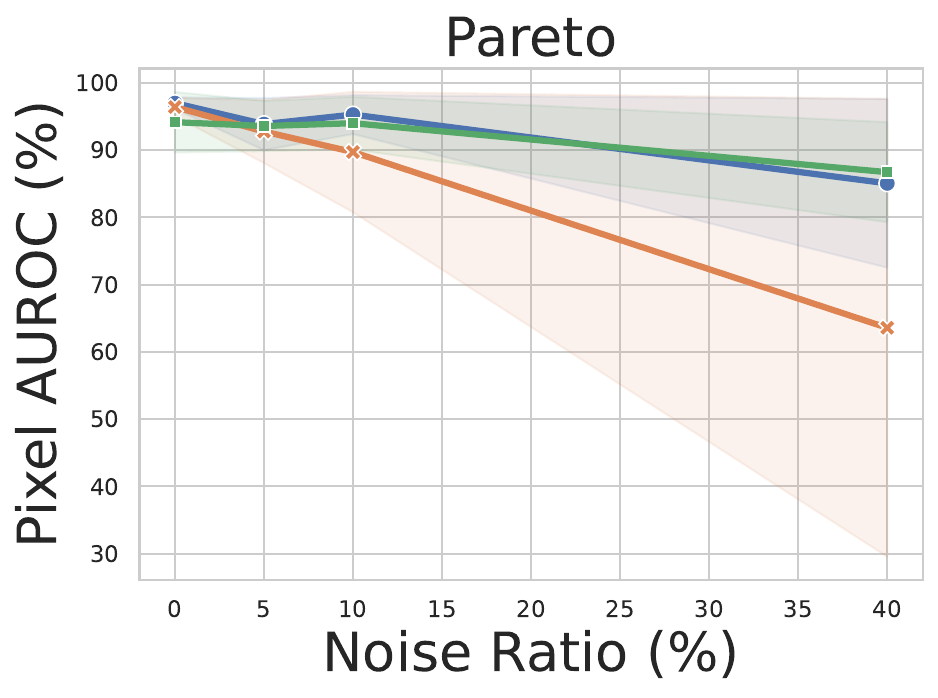}
\caption{}
\end{subfigure}
\\
\begin{subfigure}{0.49\linewidth}
\includegraphics[width=.99\linewidth]{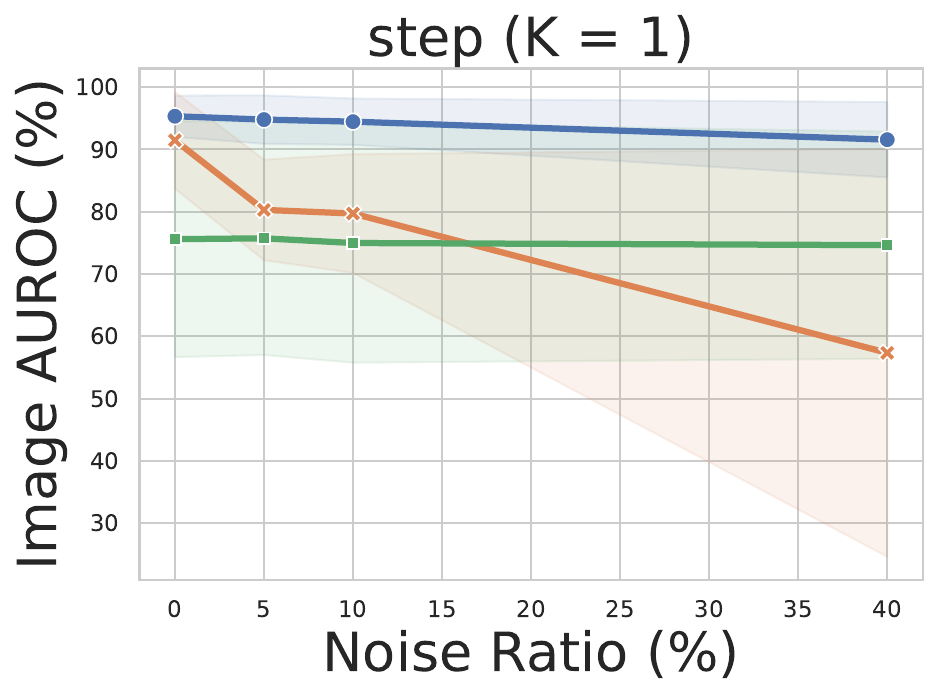}
\caption{}
\end{subfigure}
\hfill
\begin{subfigure}{0.49\linewidth}
\includegraphics[width=.99\linewidth]{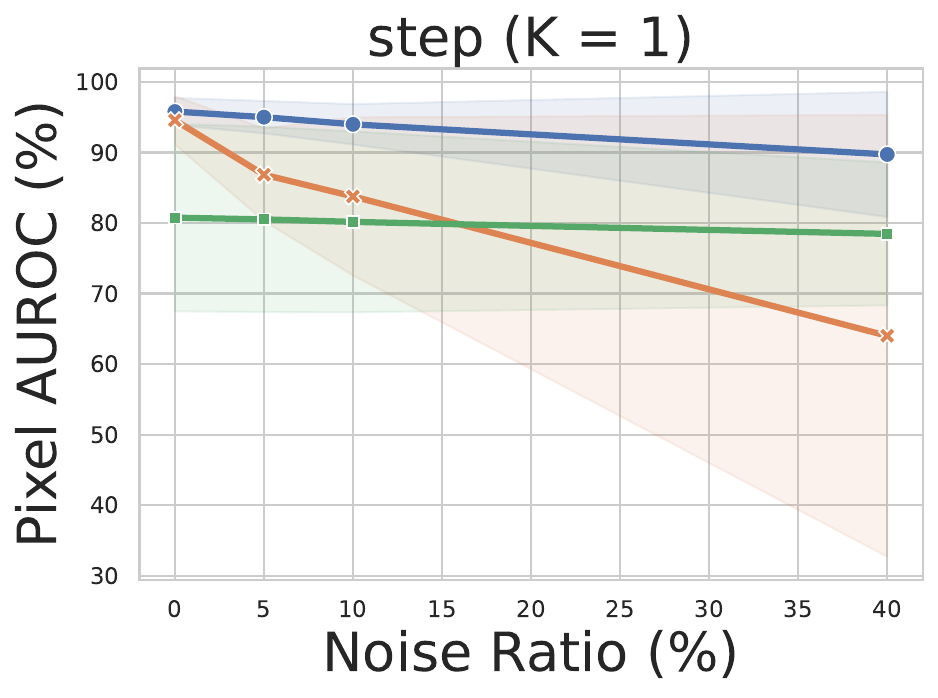}
\caption{}
\end{subfigure}
\\
\begin{subfigure}{0.49\linewidth}
\includegraphics[width=.99\linewidth]{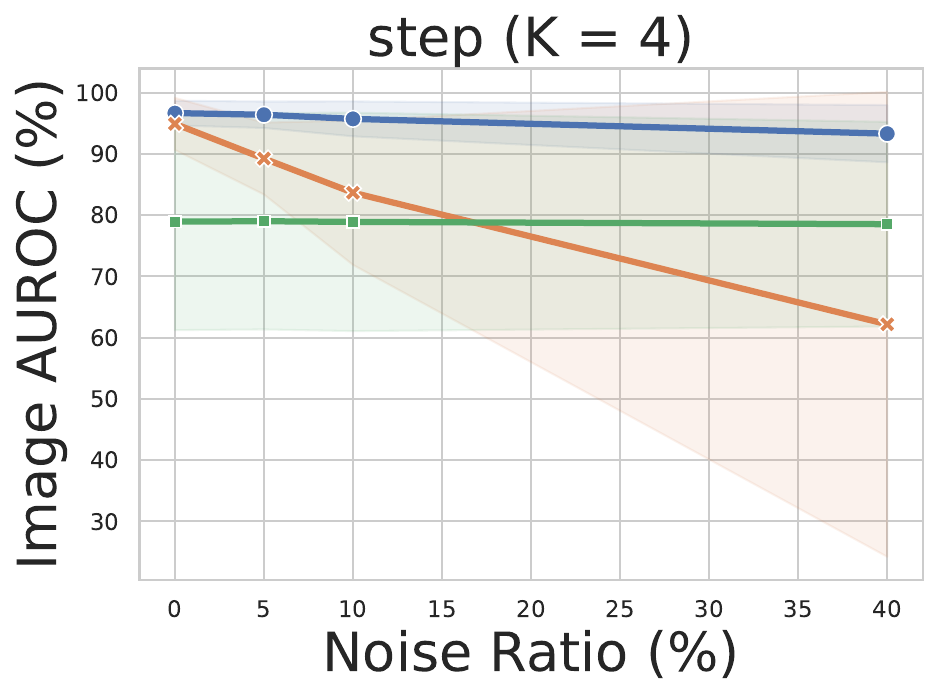}
\caption{}
\end{subfigure}
\hfill
\begin{subfigure}{0.49\linewidth}
\includegraphics[width=.99\linewidth]{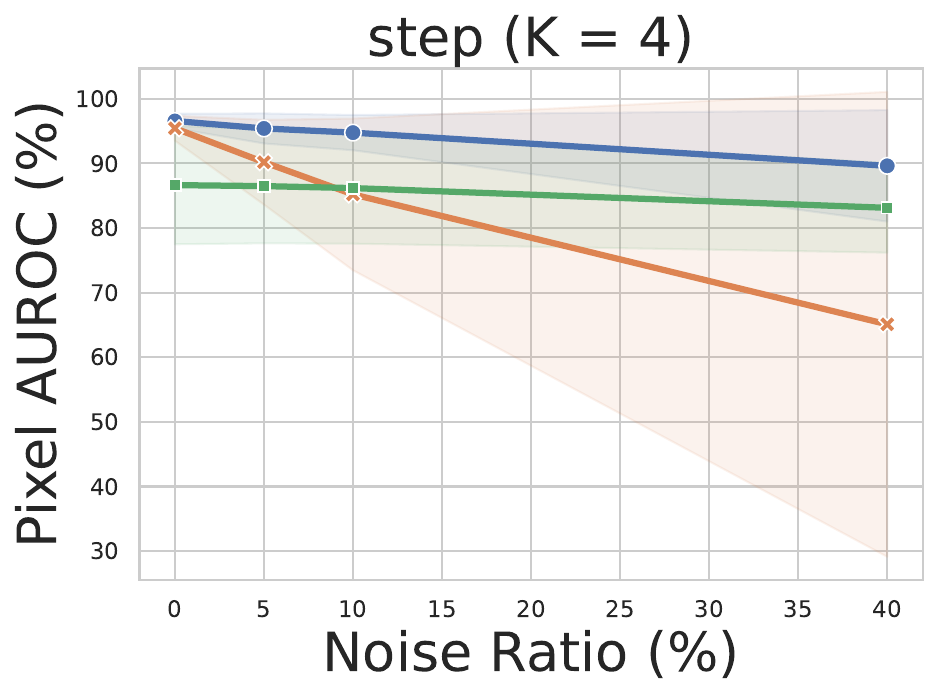}
\caption{}
\end{subfigure}
\caption{
The plot of noise ratio of train data versus anomaly classification (image-level AUROC) and segmentation (pixel-level AUROC) performance. In all cases, \ours outperforms both PatchCore and SoftPatch. Dataset is MVTecAD with different tail distributions. The lines show the mean and shades show standard deviation of multiple runs of experiments with different seeds.
}
\label{fig:noise_ratio}
\vspace{-.3cm}
\end{figure}

%KP modified table
\begin{table}[t]
\centering
\resizebox{\linewidth}{!}{
\begin{tabular}{@{}c@{}c@{\hspace{0.5em}}c@{\hspace{0.2em}}c@{\hspace{0.2em}}c@{\hspace{0.2em}}c@{\hspace{0.2em}}c@{}}
\toprule
method & \makecell{combined ratio of \\ missing $C_t$ \& \\ included AS ($\downarrow$)}  & \makecell{ratio of missing\\ $C_t$ samples ($\downarrow$)} & \makecell{ratio of \\ included AS ($\downarrow$)} & \makecell{ratio of included\\  $C_h$ samples ($\downarrow$)} & \makecell{class size \\ prediction error ($\downarrow$)} & AUROC (\%) ($\uparrow$) \\ \midrule
LOF &  89.58 $\pm$ 44.42 &  31.22 $\pm$ 34.88 &  58.36 $\pm$ 9.54 &  92.50 $\pm$ 7.87 &  1.00 $\pm$ 0.00 &  77.08 $\pm$ 25.86 \\ 
IF &  \ul{18.15 $\pm$ 15.91} &  7.25 $\pm$ 10.05 &  10.90 $\pm$ 5.86 &  69.02 $\pm$ 25.96 &  1.00 $\pm$ 0.00 &  \ul{98.96 $\pm$ 1.27} \\ 
OCSVM &  49.95 $\pm$ 9.76 &  0.78 $\pm$ 2.19 &  49.17 $\pm$ 7.57 &  95.06 $\pm$ 6.08 &  1.00 $\pm$ 0.00 &  96.72 $\pm$ 3.59 \\ 
DBSCAN &  35.36 $\pm$ 49.55 &  30.64 $\pm$ 44.38 & \bf 4.72 $\pm$ 5.17 & \bf 25.57 $\pm$ 27.24 &  \ul{0.35 $\pm$ 0.44} &  84.46 $\pm$ 22.18 \\ 
KMeans &  94.67 $\pm$ 44.95 &  47.22 $\pm$ 35.49 &  47.45 $\pm$ 9.46 &  97.40 $\pm$ 3.99 &  4.27 $\pm$ 1.40 &  52.58 $\pm$ 20.46 \\ 
GMM &  100.00 $\pm$ 0.00 &  \bf 0.00 $\pm$ 0.00 &  100.00 $\pm$ 0.00 &  97.58 $\pm$ 2.86 &  12.50 $\pm$ 1.53 &  50.00 $\pm$ 0.00 \\ 
KDE &  29.81 $\pm$ 24.15 &  2.14 $\pm$ 6.26 &  27.67 $\pm$ 17.89 &  87.47 $\pm$ 10.35 &  1.00 $\pm$ 0.00 &  97.98 $\pm$ 2.19 \\ 
AP &  27.81 $\pm$ 32.59 &  12.98 $\pm$ 24.01 &  14.83 $\pm$ 8.58 &  77.45 $\pm$ 28.21 &  0.70 $\pm$ 0.22 &  93.03 $\pm$ 12.8 \\ 
KNN &  43.49 $\pm$ 31.25 &  14.46 $\pm$ 24.07 &  29.03 $\pm$ 7.18 &  83.12 $\pm$ 18.28 &  1.00 $\pm$ 0.00 &  94.51 $\pm$ 10.34 \\ 
\sampo & \textbf{5.10 $\pm$ 5.07} &  \ul{0.12 $\pm$ 0.65} &  \ul{4.98 $\pm$ 4.42} &  \ul{31.96 $\pm$ 26.05} &  \textbf{0.19 $\pm$ 0.05} &  \textbf{99.78 $\pm$ 0.30} \\  \midrule
\multicolumn{7}{l}{\textit{Ablation on the class size predictor:}}\\  
\samp ($p$=100) &  8.48 $\pm$ 10.88 &  1.07 $\pm$ 5.23 &  7.41 $\pm$ 5.65 &  42.64 $\pm$ 31.86 &  0.51 $\pm$ 0.04 &  99.76 $\pm$ 0.29 \\ 
\samp ($p$=50) &  7.37 $\pm$ 10.35 &  3.46 $\pm$ 6.80 &  3.91 $\pm$ 3.55 &  27.90 $\pm$ 27.36 &  0.12 $\pm$ 0.12 &  97.36 $\pm$ 5.65 \\ 
\samp (top-1) &  14.47 $\pm$ 7.99 &  0.20 $\pm$ 0.78 &  14.27 $\pm$ 7.21 &  73.97 $\pm$ 27.95 &  0.85 $\pm$ 0.04 &  99.73 $\pm$ 0.34 \\ 
\samp (average) &  6.48 $\pm$ 10.11 &  3.65 $\pm$ 6.99 &  2.83 $\pm$ 3.12 &  22.16 $\pm$ 24.33 &  0.21 $\pm$ 0.07 &  99.35 $\pm$ 2.59 \\
\bottomrule
\end{tabular}
}
\caption{The analysis on the proposed \samp and its class size predictor. The results are obtained from both MVTecAD and VisA and averaged across all different types of tail distributions and random seeds (overall 60 different seeds) with the standard deviations indicated on the right. The best and second best performances are marked in \textbf{bold} and \ul{underline}, respectively. Acronyms/notations: IF: Isolation Forest; AP: Affinity Propagation; AS: anomaly samples; $C_h$ / $C_t$: head / tail classes.
%\KP{I see. Please explain why that metric is the lower the better in the main text to help the readers understand.}
}
\label{table:analysis}
% \vspace{-.2cm}
\end{table}

%KP modified table
\begin{table}[t]
\centering
\resizebox{\linewidth}{!}{
\begin{tabular}{@{}c@{\hspace{0.6em}}c@{\hspace{0.6em}}c@{\hspace{0.6em}}c@{\hspace{0.6em}}cc@{\hspace{0.6em}}c@{\hspace{0.6em}}cc@{\hspace{0.6em}}c@{\hspace{0.6em}}c@{}}
\toprule
\multirow{2.5}{*}{task} &tail type &\multicolumn{3}{c}{Pareto} &\multicolumn{3}{c}{step ($K=4$)} &\multicolumn{3}{c}{step ($K=1$)}\\
\cmidrule(r){3-5} \cmidrule(r){6-8} \cmidrule(r){9-11}
 &class type &$C_t$ &$C_h$ &all &$C_t$ &$C_h$ &all &$C_t$ &$C_h$ &all\\
\midrule
\multirow{2}{*}{\shortstack{anomaly\\classification}} &w/o FSA &82.22 &91.22 &88.13 &60.01 &\bf 97.60 &75.05 &66.28 &\bf 97.85 &78.94\\
 &w/ FSA &\bf 96.55 &\bf 95.24 &\bf 96.13 &\bf 93.54 & 95.78 &\bf 94.44 &\bf 95.82 & 95.34 &\bf 95.72\\
\midrule
\multirow{2}{*}{\shortstack{anomaly\\segmentation}} &w/o FSA &91.60 &94.94 &93.54 &68.99 &\bf 96.85 &80.14 &79.39 &\bf 96.35 &86.23\\
 &w/ FSA &\bf 96.09 &\bf 95.01 &\bf 95.30 &\bf 94.19 &93.71 &\bf 94.00 &\bf 95.56 &93.20 &\bf 94.75\\
\bottomrule
\end{tabular}
}
\caption{Abaltion on the few-shot augmentation (FSA) of \ours. The image-level and pixel-lavel AUROC metrics (\%) are reported on the MVTecAD dataset with different types of tail distributions for anomaly classification and segmentation tasks, respectively. Notations: $C_h$ / $C_t$: head / tail classes. 
}
\label{table:ablation_aug}
\vspace{-.2cm}
\end{table}

\begin{figure*}[t]
\centering
\begin{subfigure}{0.24\linewidth}
\centering
\includegraphics[width=.99\linewidth]{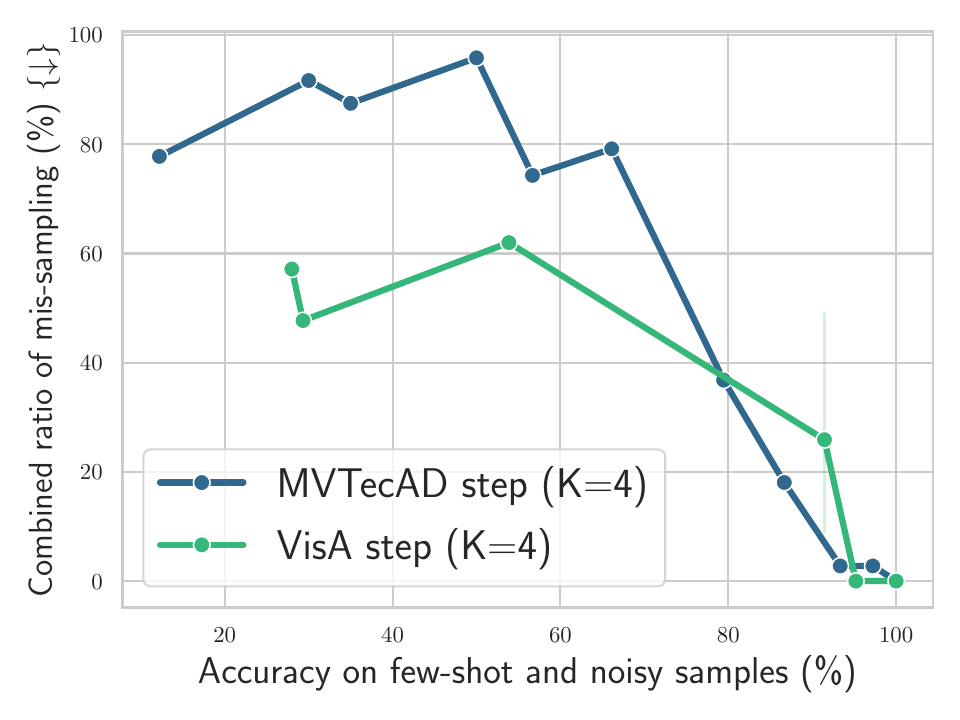}
\caption{}
\end{subfigure}
\hfill
\begin{subfigure}{0.24\linewidth}
\includegraphics[width=.99\linewidth]{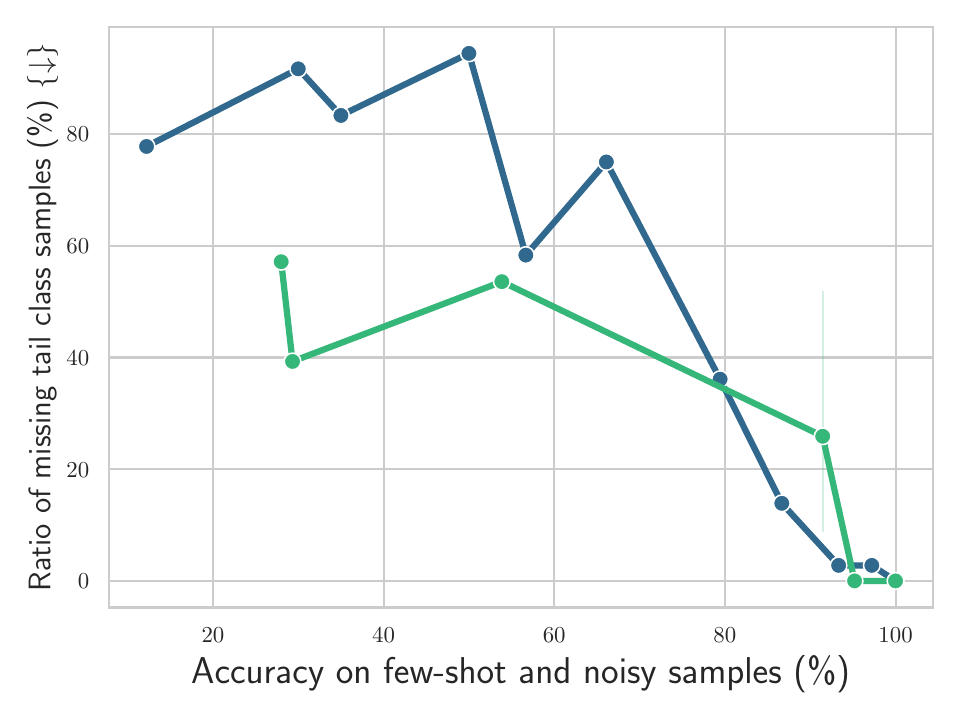}
\caption{}
\end{subfigure}
\hfill
\begin{subfigure}{0.24\linewidth}
\includegraphics[width=.99\linewidth]{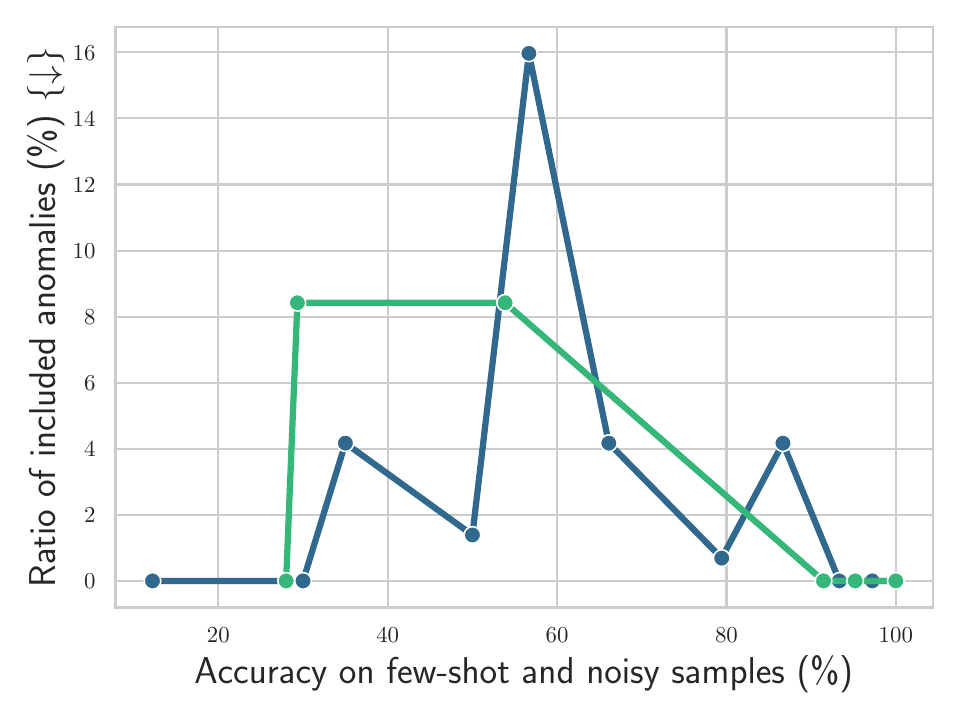}
\caption{}
\end{subfigure}
\hfill
\begin{subfigure}{0.24\linewidth}
\centering
\includegraphics[width=.99\linewidth]{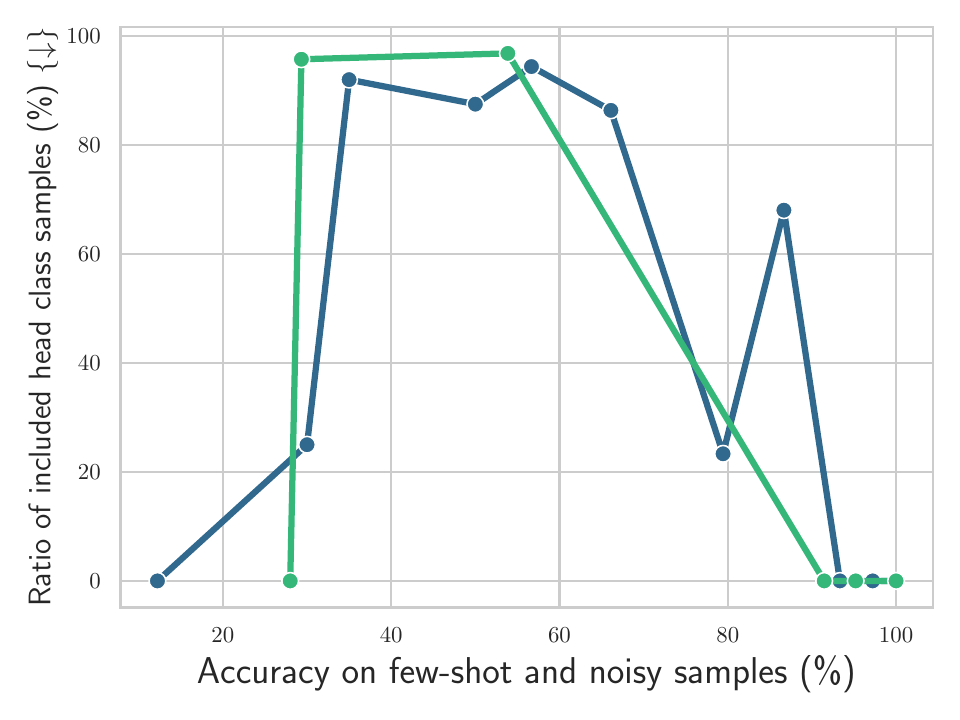}
\caption{}
\end{subfigure}
\\
\begin{subfigure}{0.24\linewidth}
\includegraphics[width=.99\linewidth]{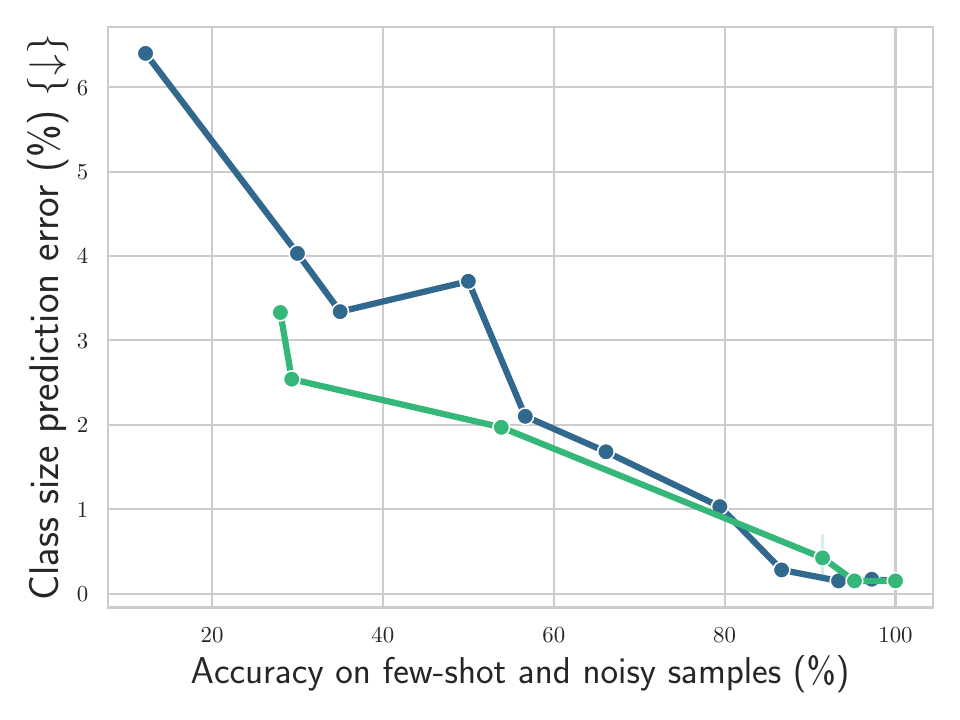}
\caption{}
\end{subfigure}
\hfill
\begin{subfigure}{0.24\linewidth}
\includegraphics[width=.99\linewidth]{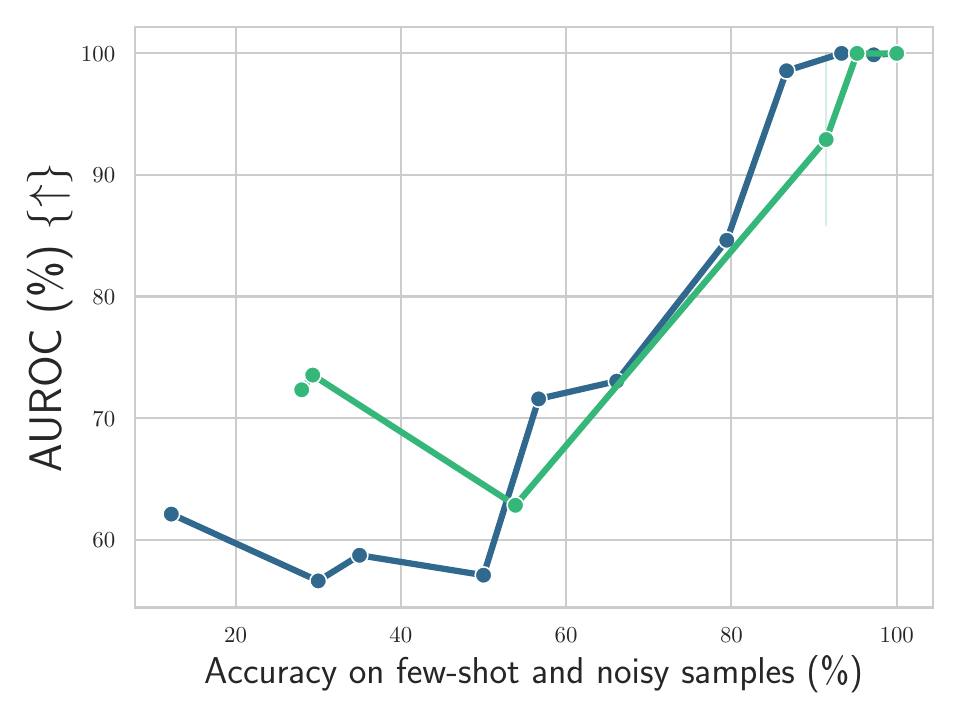}
\caption{}
\end{subfigure}
\hfill
\begin{subfigure}{0.24\linewidth}
\includegraphics[width=.99\linewidth]{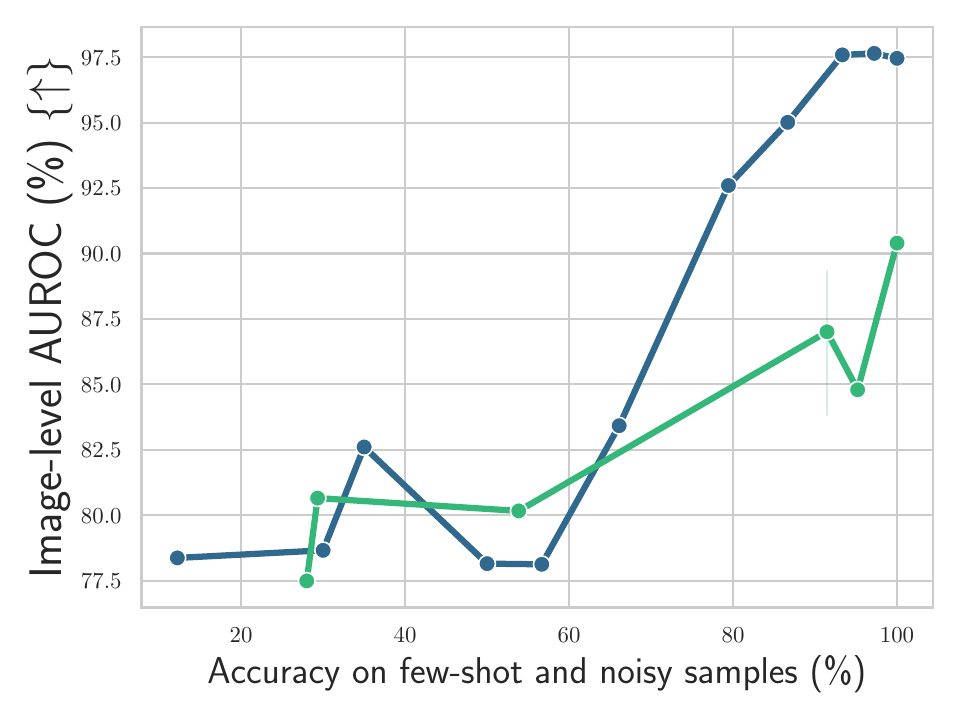}
\caption{}
\end{subfigure}
\hfill
\begin{subfigure}{0.24\linewidth}
\includegraphics[width=.99\linewidth]{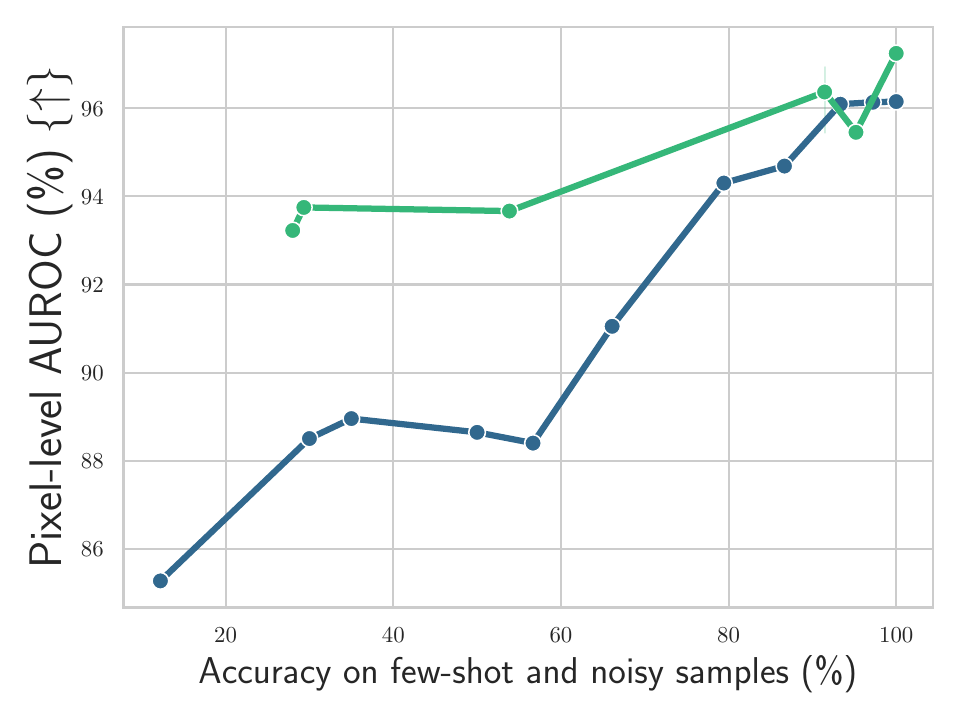}
\caption{}
\end{subfigure}
\caption{
Classification accuracy between tail classes and noisy samples versus metrics relevant to class size prediction and few-shot sampling on MVTecAD and VisA datasets with step-like tailed distribution and $K=4$. The correlation is strong for (a) mis-sampling ratio, (b) ratio of missing few-shot samples, (e) class size prediction error, and (f) AUROC for predicting whether a sample is few-shot or not. We show that improving the discriminative aspect of network's embeddings improves few-shot sampling and prediction error of class size predictor, which in turn improves (g) anomaly classification (image-level AUROC) and (h) anomaly segmentation (pixel-level AUROC) performance  of \ours. 
}
\label{fig:acc_vs}
\vspace{-.2cm}
\end{figure*}

\subsubsection{Class size prediction}
We measure how accurately our class size predictor estimates the class size of each sample by measuring the class size prediction error. It is measured for each train sample and weight-averaged through the whole train samples:
\begin{equation}
error
=
\frac{1}{| \mathcal{X} |}
w_i
\sum_{i=1}^{|\mathcal{X}|}
\left|\kappa_i - |C_{k(i)}| \right|
\end{equation}
where $k(i)$ indicates the index such that $x_i \in C_{k(i)}$ (namely, $ C_{k(i)}$ is the class of $x_i$). The weight is chosen by $w_i = 1 / |C_{k(i)}|$, which is to give more penalty on the few-shot class samples more than the head class samples.

Tab.~\ref{table:analysis} shows that most of the clustering based methods vastly fail on this task, and the estimation by our class size predictor outperforms them significantly. We believe this is partly due to suboptimal hyperparameters used for these algorithms. Due to their classical nature, however, it is unclear how to tune them without the availability of the validation set in our scenario.

\subsubsection{Abalation study on class size predictor}
We conduct detailed ablation study on the proposed class size prediction in Eq.~\eqref{eq:cize}. 
One hyperparameter used in \samp $S_{tail}$ of \ours is the percentiling proportion $p$ to consider only the majority of nearby samples within the neighborhood. We analyze how this parameter can affect the anomaly detection performance of \ours. Tab.~\ref{table:analysis} indicates that using the nearest 85\% embeddings (\ie, $p=85\%$) is the best choice. Using all of the available neighborhoods (\ie, $p=100\%$) would include different class samples in the neighborhood, deterioriating class estimation. On the other hand, counting only with the half of samples (\ie, $p=50\%$) in the neighborhood may lose too much information.

Another component in the class size predictor that requires analysis is its voting mechanism used in Eq.~\eqref{eq:cize}. One may simply use the nearest embedding's neighborhood size instead, or the class size of $x_i$ can be estimated by averaging instead of majority vote. As indicated in Tab.~\ref{table:analysis}, both the nearest embedding (\ie, top-1) and averaging are inferior to majority voting. The former is inevitably less robust, while averaging can still be sensitive to the outlier values of neighborhood size. 

\subsubsection{Ablation on \ours}

We analyze the impact of augmentation of few-shot class patches by \samp in \ours. The \ours model without these augmented patches has the same memory bank as SoftPatch, but its inference scoring is simpler. Tab.~\ref{table:ablation_aug} indicates that without the augmented patches from \samp, the memory-based model cannot detect anomalies in few-shot class samples. This shows the necessity of tail class augmentation by \samp. We note some performance decrease in the head class detection performance since few-shot augmentation may involve few anomaly samples from the head classes that behave like few-shot class instances.  This issue intrinsically is due to the weakness of encoder embedding, and the proposed \samp based on such embeddings can induce incorrectly sampled few-shot class instances, which we discuss in Sec.~\ref{sec:limit}. 

\subsection{Noise ratio}
We analyze the impact of noise ratio. Training data with a greater number of defect samples increase the chance of sampling noise instances for the few-shot sampler. We analyze this trend in Fig.~\ref{fig:noise_ratio}, but the impact is not severe. In all cases, \ours either outperforms or is on par with both PatchCore and SoftPatch.

\subsection{Relation between the discriminative aspect of embedding and \ours}
\ours utilizes the embedding features $e_i$ of training samples $x_i$ for few-shot sampling and augmentation thereof. Therefore, \ours's detection performance highly relies on the discriminative aspect of the embedding feature. Poorly designed embedding features will not discriminate different class samples in the embedding space. In this case, \samp cannot sample tail class samples and it may overly many head class samples. \ours's memory bank obtained so will neither be class-balanced nor noise-free, making the inference of \ours poor.

In contrast, a highly discriminative embedding features will make \ours's class size prediction accurate, thereby enabling it to sample the few-shot class samples exclusively without aggregating any anomaly and head class samples. In which case, \ours is both noise-free and class-balanced.

We test this behavior by Oracle labeling and training the embedding features with class labels. Fig.~\ref{fig:acc_vs} shows that improving discriminative quality of embedding (measured by its classification accuracy on noise and tail class samples) accordingly improves the relevant metrics. Particularly, more discriminative embeddings make more accurate class size prediction, improving \samp along with its anomaly classification and segmentation performances.

\section{Limitations}
\label{sec:limit}
The class size prediction in \samp can fail if the reflective-symmetric assumption on the inter-class and intra-class similarities breaks down. Particularly, this case can occur if the utilized encoder embedding is either poor or not aligned with the label space of classes. In some cases, the geometric aspect of defect samples highly resembles that of few-shot class instances in the embedding space. We believe that this is due to imperfect nature of encoder embedding, which is often more sensitive to spurious features than the notion of class and object-centric aspect \cite{ming2022impact}. This issue can be improved by enhancing the encoder network.

\section{Conclusion}
In this study, we introduced \ours, a novel memory-based anomaly detection model designed to effectively address the challenges of unsupervised anomaly detection in noisy and long-tail class distributions. By developing a unique class size predictor and a tailored memory bank, we successfully navigated the tail-versus-noise dilemma, enhancing the model's performance in identifying anomalies within a contaminated and imbalanced dataset. Our comprehensive evaluations demonstrated \ours's superior ability to mitigate noise contamination and class imbalance, underscoring its significance in advancing anomaly detection technologies. This work not only showcases a significant stride in handling real-world anomaly detection scenarios but also lays a robust foundation for future research in this critical field.

\paragraph{Acknowledgments}
\thanks{Yoon Gyo Jung and Octavia Camps were supported by NSF grant 2038493, ONR grant N00014-21-1-2431, NIH grant R01CA240771 from NCI, and U.S. Department of Homeland Security grant 22STESE00001-03-02. Jaewoo Park and Wonchul Kim were supported by the Technology Innovation Program (20023364) funded by the Ministry of Trade, Industry \& Energy (MOTIE, Korea). Kuan-Chuan Peng was exclusively supported by Mitsubishi Electric Research Laboratories. The views and conclusions contained in this document are those of the authors and should not be interpreted as necessarily representing the official policies, either expressed or implied, of the U.S. Department of Homeland Security.}

{
\small
\bibliographystyle{ieeenat_fullname}
\bibliography{main}
}

\clearpage
\setcounter{page}{1}
\maketitlesupplementary

\section{Supplementary to Method}
\label{supp_sec:theory}
\setcounter{thm}{0}

% The following proposition shows that the radius of neighborhood given by Eq.~\eqref{eq:radius} corresponds to the decision boundary that maximizes the inter-class separation based on the following assumption and proposition.
% \vspace{-.15cm}
% \begin{ass*}
% \label{ass:symmetry}
% Let $p_{same}( \cdot | e_i)$ and $p_{diff}( \cdot | e_i)$ denote, respectively, the distributions of embedding similarities of the same and different classes. Then, $p_{same}(s | e_i) = p_{diff}(-s | e_i)$. Fig.~\ref{fig:proof_emp} in Supp.~\ref{supp_sec:theory} shows the feasibility of such symmetric assumption.
% %\begin{equation}
% %p_{same}(s | e_i) = p_{diff}(-s | e_i).
% %\end{equation}
% \end{ass*}

% % TODO: revision is required
% \vspace{-.15cm}
% \begin{prop}
% \label{prop:opt_th}
% The decision boundary that maximizes the separation between $p_{same}(\cdot | e_i)$ and  $p_{diff}(\cdot | e_i)$ corresponds to $r_i = 1- \cos (\arccos (m_i) / 2) $. Proof is given in Supp.~\ref{supp_sec:theory}
% \end{prop}

\begin{figure*}[t!]
\centering
\begin{subfigure}{0.275\linewidth}
\centering
\includegraphics[width=.99\linewidth]{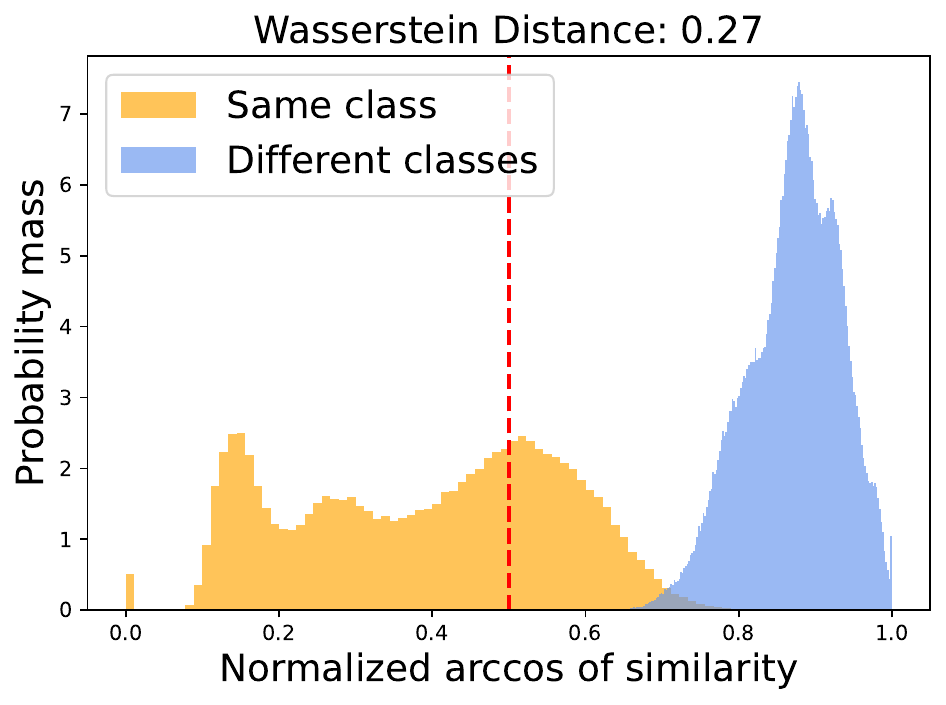}
\caption{MVTecAD $K=1$}
\end{subfigure}
\begin{subfigure}{0.275\linewidth}
\includegraphics[width=.99\linewidth]{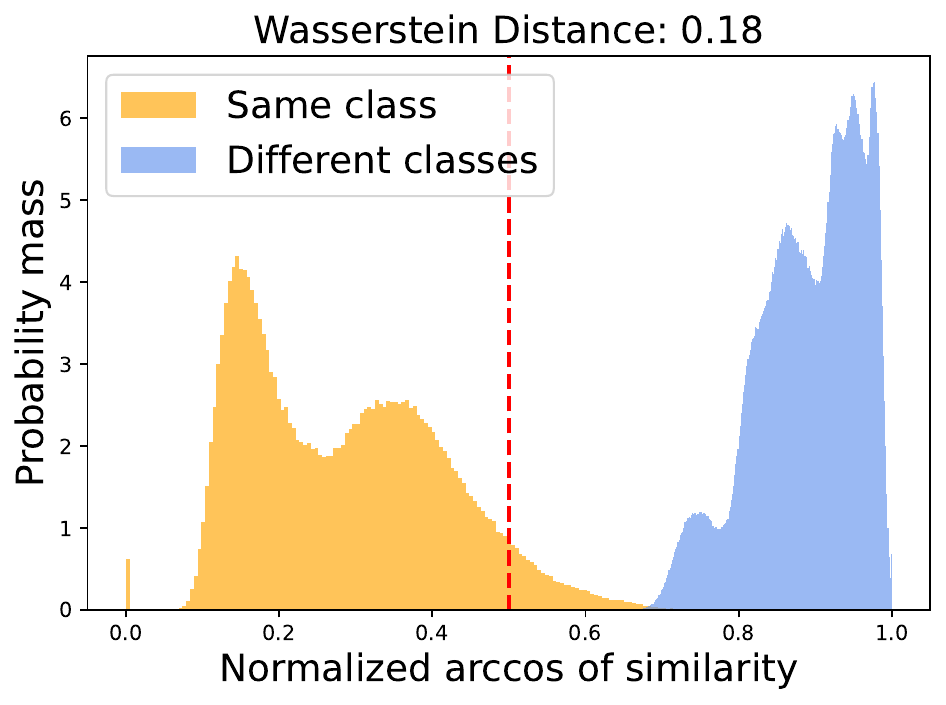}
\caption{MVTecAD $K=4$}
\end{subfigure}
\begin{subfigure}{0.275\linewidth}
\includegraphics[width=.99\linewidth]{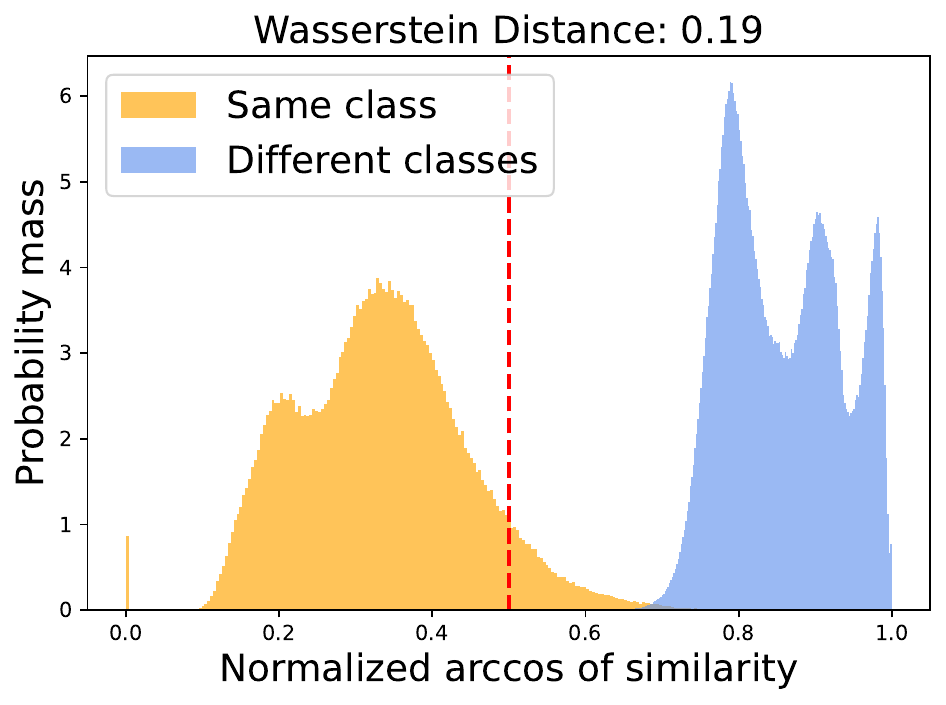}
\caption{MVTecAD Pareto}
\end{subfigure}
\\
\begin{subfigure}{0.275\linewidth}
\centering
\includegraphics[width=.99\linewidth]{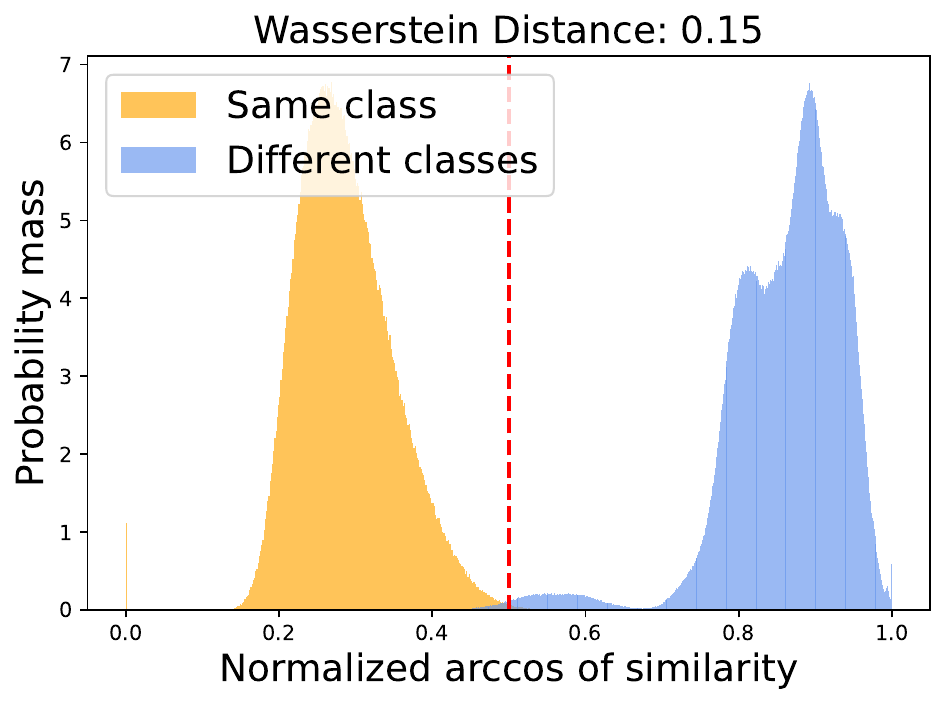}
\caption{VisA $K=1$}
\end{subfigure}
\begin{subfigure}{0.275\linewidth}
\includegraphics[width=.99\linewidth]{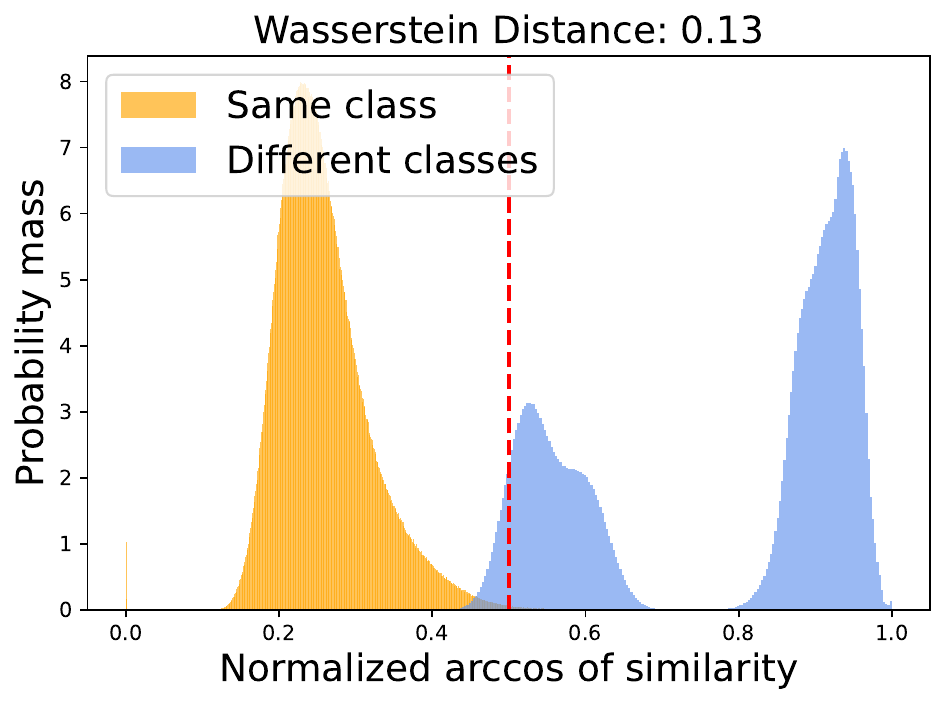}
\caption{VisA $K=4$}
\end{subfigure}
\begin{subfigure}{0.275\linewidth}
\includegraphics[width=.99\linewidth]{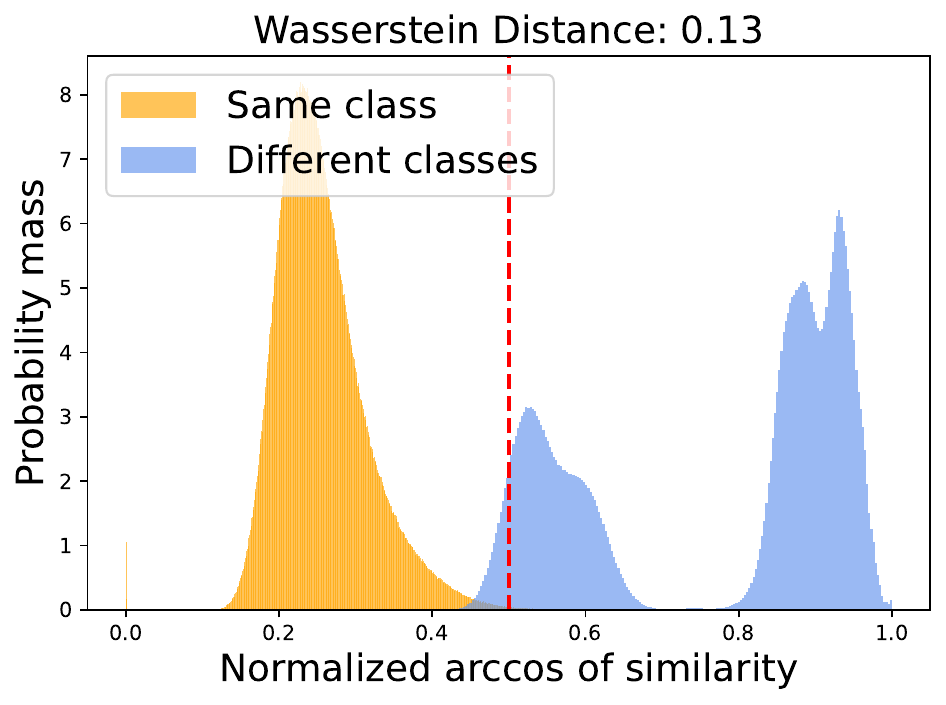}
\caption{VisA Pareto}
\end{subfigure}
\caption{Empirical validation of the assumption in Sec.~\ref{sec:method}, showing that the inter-class and intra-class similarity distributions approximately exhibit reflective symmetry discussed in the assumption in Sec.~\ref{sec:method} and \ref{supp_sec:theory}} 
\label{fig:proof_emp}
\end{figure*}

We state the assumption and proposition from Sec.~\ref{sec:method} here in a more rigorous, detailed manner:

\begin{ass*}
\label{ass:supp_symmetry}
Let $p_{same}( s | e_i)$ denote the distribution of cosine similarity between $e_i$ and (random vector) embedding $e$ of a same class sample. On the other hand, let $p_{diff}( s | e_i)$ denote the distribution of cosine similarity between $e_i$ and (random vector) embedding $e$ of a different class sample. Then, a reflective symmetry holds between $p_{same}$ and $p_{diff}$ in terms of their angles:
\begin{equation}
p_{same}( \cos(a) | e_i) = p_{diff}( \cos(a_{\max} - a) | e_i).
\end{equation}
where $a = \arccos(s)$ and $a_{\max} = \arccos(m_i)$. Recall that $m_i$ is the minimum of the support of the inter-class similarity distribution $p_{diff}(\cdot | e_i)$.
\end{ass*}
Fig.~\ref{fig:proof_emp} in Supp.~\ref{supp_sec:theory} shows the feasibility of the above symmetric assumption.
We prove Prop.~\ref{prop:opt_th} under a reasonable regularity condition, whose trend is evidenced by Fig.~\ref{fig:proof_emp}. Below, we denote by $supp$ the support of a distribution, and for notational brevity, we let $supp(p_{same}) := supp (p ( \cdot | e_i ) )$ and similarly for $p_{diff}$.

\begin{prop}
\label{prop:opt_th}
For distributions $p_{same}(s | e_i)$ and $p_{diff}(s | e_i)$ that are weakly increasing and decreasing, respectively, let $supp(p_{same})$ be disjoint with $ supp(p_{diff})$ and $\max supp( p_{same}) = 1$ Then the threshold given by $\tau_i = \cos (\arccos (m_i) / 2) $ separates the supports by
\begin{equation}
\max supp (p_{diff})
<
\tau_i
<
\min supp (p_{same}),
\end{equation}
and maximizes the distance between $\tau_i$ and  $supp (p_{diff}) \cup supp (p_{same})$ under arccosine transformation.
\end{prop}

Note that the radius satisfies $r_i = 1 - \tau_i$

\begin{proof}
Let $f(a) = p_{same}( \cos a)$ and $g(a) = p_{diff}( \cos a)$ for $0 \leq a \leq \pi$. Then, by the assumptions from the above, $f(a)$ is decreasing and $g(a)$ is increasing. Since  $\max supp( p_{same}) = 1$, we have $\min supp f = 0$. Due to the reflective symmetry, we have
\begin{equation}
f(a) = g(a_{\max} - a),
\end{equation}
or
\begin{equation}
f(a_{\max}/2 - a) = g(a_{\max}/2 + a).
\end{equation}
Since $\cos$ is 1-to-1 for the given range of $a$, disjointness of supports for $p_{same}$ and $p_{diff}$ is preserved for the supports of $f$ and $g$. Since $f$ is decreasing and $g$ is increasing, 
\begin{equation}
a_{\max} /2 \notin supp(f) \cup supp(g),
\end{equation}
showing that $a_{\max} / 2$ separates the supports of $f$ and $g$. Moreover, 
\begin{equation}
a_{\max} / 2 =  \underset{a}{\arg\max} \, d(a, supp(f) \cup supp(g)))
\end{equation}
where $d$ is set distance. Observing that
\begin{equation}
a_{\max} /2 = \arccos(m_i) / 2 = \arccos(r_i)
\end{equation}
completes the proof.
\end{proof}

\section{Supplementary to Analysis}
\label{supp_sec:analysis}

\begin{figure}[t]
\centering
\includegraphics[width=.4\linewidth]{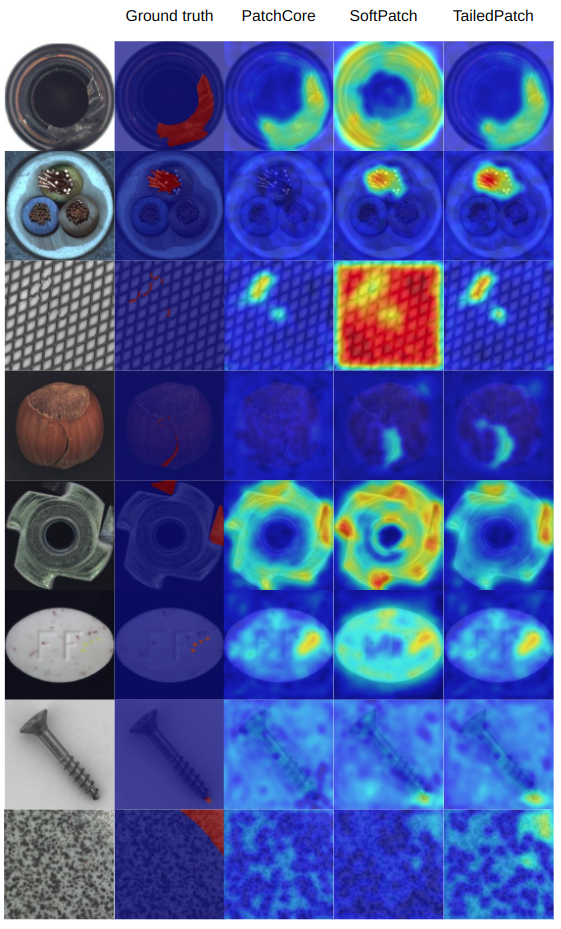}
\caption{
Qualitative analysis of  PatchCore, SoftPatch, and \ourso.
}
\label{fig:supp_qual}
\end{figure}

\subsection{Qualitative analysis}

Qualitative results of segmentation heatmap are given in Fig. ~\ref{fig:supp_qual}, which shores the anomaly scores on the image pixel level. The result shows a trend that \ours resolves a performance trade-off between SoftPatch and PatchCore.

\begin{table*}[t]
\caption{Experiments on \ours with mix perturbed data in the tail class denoted as TailPert where $p$ is the perturbed ratio, SoftPatch + DBSCAN/K-Means, using transformer's classification token for TailSampler, and noise in tail classes denoted as NoisyTail. 
}
\label{table:rebuttal}
\centering
\resizebox{\linewidth}{!}{
\begin{tabular}{r@{\hspace{0.6em}}c@{\hspace{0.6em}}c@{\hspace{0.6em}}cc@{\hspace{0.6em}}c@{\hspace{0.6em}}cc@{\hspace{0.6em}}c@{\hspace{0.6em}}c}
\toprule
tail type &\multicolumn{3}{c}{Pareto} &\multicolumn{3}{c}{step ($K=4$)} &\multicolumn{3}{c}{step ($K=1$)}\\
\cmidrule(r){2-4} \cmidrule(r){5-7} \cmidrule(r){8-10}
class type &$C_t$ &$C_h$ &all &$C_t$ &$C_h$ &all &$C_t$ &$C_h$ &all\\
\midrule
TailPert $p=0.1$ & 96.4 / 96.2 & 95.2 / 94.1 & 95.9 / 95.2 & 96.1 / 96.1 & 95.9 / 93.5 & 96 / 95.1 & 93.7 / 93.8 & 95.4 / 92 & 94.4 / 93.1 \\ 
TailPert $p=0.3$ & 96.2 / 96 & 95.1 / 94.1 & 95.8 / 95.1 & 96.1 / 96.1 & 95.9 / 93.5 & 96 / 95.1 & 93.5 / 93.7 & 95.7 / 91.9 & 94.4 / 93 \\ 
TailPert $p=0.5$ & 96 / 96 & 94.5 / 94.9 & 95.8 / 95.1 & 95.9 / 96.1 & 95.6 / 93 & 95.8 / 94.9 & 92.5 / 93.3 & 95.5 / 92 & 93.7 / 92.8 \\ 
\midrule
DBSCAN \cite{ester1996density}-SoftPatch \cite{jiang2022softpatch} & 87 / 93.5 & 93.3 / 95.3 & 90.6 / 94.4 & 71.8 / 84.2 & 97.6 / 96.3 & 82.2 / 89.1 & 63.5 / 71.3 & 97.8 / 96.5 & 77.2 / 81.4 \\ 
Kmeans \cite{arthur2007k}-SoftPatch \cite{jiang2022softpatch} & 86.7 / 92.8 & 92.5 / 95.1 & 90.5 / 94.1 & 71.8 / 83.4 & 97.6 / 96.4 & 82.2 / 88.7 & 63.5 / 71 & 97.8 / 96.8 & 77.2 / 81.3 \\ 
\midrule
ViT-L \cite{dosovitskiy2020image} ClsToken & 96.6 / 96.5 & 95.3 / 95.3 & 96.1 / 95.7 & 96 / 95.5 & 95.9 / 94.4 & 96 / 95.2 & 93.4 / 94.2 & 96.1 / 95.1 & 94.5 / 94.6 \\ 
\midrule
Noisy Tail & - & - & - & 92.3 / 94 & 95.6 / 92.4 & 93.7 / 93.5 & - & - & - \\ 
\midrule
SoftPatch \cite{jiang2022softpatch} & 84.7 / 92.2 & 87.0 / 93.8  & 87.7 / 93.41 & 67.7 / 81  & 97.5 / 96.5  & 79.6 / 87.2  & 60.7 / 70.3  & 97.5 / 96.9  & 75.4 / 81  \\
\midrule
Ours & 96.6 / 96.1 & 95.2 / 95 & 96.1 / 95.3 & 95.8 / 95.6 & 95.3 / 93.2 & 95.7 / 94.7 & 93.5 / 94.2 & 95.8 / 93.7 & 94.4 / 94 \\
% TailPerturb $p=0.7$ & 95.77 & 95.15 & 95.78 & 94.54 & 95.50 & 95.00 & 90.85 & 95.90 & 92.87 \\
\bottomrule
\end{tabular}
}
\end{table*}

\subsection{Additional experiments and analysis}
We conduct additional experiments in order to verify the strengthness of \ours with results in Tab.~\ref{table:rebuttal}. First of all, we perturb the tail class samples with $brightness=0.5, constrast=0.5, saturation=0.5, hue=0.1$ perturbation with the ratio $p \in \{0.1, 0.3, 0.5\}$. Besides the extreme case where $k=1$, the effect of perturbation is not remarkable showing the robustness of \ours. 

Secondly, we verify the robustness of TailSampler by substituting it with other unsupervised clustering methods which are DBSCAN \cite{ester1996density} and K-Means \cite{arthur2007k} clustering. For this experiment, we have set the threshold of tail classes to less than 20 samples and normalized with L2 normalization. For hyperparameters, we used $eps=0.5$, $min\_samples=2$ for DBSCAN and $k=20$ for KMeans. DBSCAN-SoftPatch sometimes successfully generated 15 clusters+noise class resulting in 16 clusters in total, but often fails and generates more or less clusters. For Kmeans-SoftPatch, we have tried $k=15$, which is the number of classes of MVTec-AD, but this failed detecting any long-tailed clusters. Therefore, we attempted $k=20$ and acquired results as shown in Tab.~\ref{table:rebuttal}.

Additionally, we substitute CNN with ViT-L \cite{dosovitskiy2020image}, specifically the global average pooling feature of CNN with classification token for TailSampler which increased the results slightly as shown in Tab.~\ref{table:rebuttal} denoted as ”ViT-L ClsToken”.

We also test our method when tail classes are contaminated with noise denoted as Noisy Tail. Without the information of classes in an anomalous few-shot scenario, anomalies can't be statistically classified as abnormal from a statistical perspective. Our method tends to specifically address only those instances that are statistically confirmed to be anomalous. Such scenario is tested by including one noisy samples in each of the tail classes with $K=4$ setup (1 noisy sample and 3 normal samples for each tail classes) and the results are shown in Tab.~\ref{table:rebuttal}. The results show a minor decrease in performance in tail classes, however, the mechanism of our method is fundamentally unable to address such scenarios well.

Finally, we have done experiments on MVTec-Loco \cite{bergmann2022beyond} dataset to test logical anomaly on PatchCore and our method where results are in Tab.~\ref{table:loco}. Due to the noisy head class and few-shot tail classes both of the methods do not work well, however, our method still shows better performance comparing with PatchCore.

\begin{table*}[t]
\caption{MVTec-LOCO image AUROC (\%) with (all / structural / logical) format.
}
\label{table:loco}
\centering
\resizebox{\linewidth}{!}{
\begin{tabular}{r@{\hspace{0.6em}}c@{\hspace{0.6em}}c@{\hspace{0.6em}}cc@{\hspace{0.6em}}c@{\hspace{0.6em}}c}
\toprule
tail type &\multicolumn{3}{c}{Pareto} &\multicolumn{3}{c}{step ($K=4$)}\\
\cmidrule(r){2-4} \cmidrule(r){5-7} 
class type &$C_t$ &$C_h$ &all &$C_t$ &$C_h$ &all\\
\midrule
Patchcore & 60.7 / 62.8 / 59.1 & 59.4 / 76.5 / 45.4 & 60.2 / 68.3 / 53.6 & 61.7 / 71.1 / 55.2 & 63.8 / 70.9 / 59.4 & 62.5 / 71 / 56.9 \\
SoftPatch & 52.3 / 57.8 / 52.9 & 66.5 / 67.8 / 74.5 & 58 / 61.8 / 61.6 & 62.7 / 60.3 / 68 & 61.7 / 68.4 / 64.3 & 62.3 / 63.5 / 66.5 \\
Ours & 63 / 68 / 59.4 & 66 / 74.8 / 59.3 & 64.2 / 70.7 / 59.4 & 65.9 / 71.7 / 61.7 & 67.1 / 68.8 / 66.6 & 66.4 / 70.6 / 63.6  \\
\bottomrule
\end{tabular}
}
\end{table*}

\section{Supplementary to Dataset}
\label{supp_sec:dataset}
\paragraph{\textbf{Number of total and anomaly samples}} We provide the detailed number of total and anomaly samples used for our experiments in Tabs. ~\ref{table:supp_num_mvtec_pareto}, ~\ref{table:supp_num_mvtec_k4}, ~\ref{table:supp_num_mvtec_k1}, ~\ref{table:supp_num_visa_pareto}, ~\ref{table:supp_num_visa_k4}, and ~\ref{table:supp_num_visa_k1}.
The names of unuspervised long-tail noisy anomaly detection datasets are summarized as follows:
\begin{itemize}
\item \textbf{\textit{MVTecAD-Pareto}}: MVTecAD dataset where its class distribution follows Pareto distribution and 10\% of train samples are noises in the head classes
\item \textbf{\textit{MVTecAD-step-K4}}: MVTecAD dataset where its class distribution is imbalanced such that 9 out of 15 classes have only 4 samples for each class and 10\% of train samples are noises in the head classes
\item \textbf{\textit{MVTecAD-step-K1}}: MVTecAD dataset where its class distribution is imbalanced such that 9 out of 15 classes have only 1 samples for each class and 10\% of train samples are noises in the head classes
\item \textbf{\textit{VisA-Pareto}}: VisA dataset where its class distribution follows Pareto distribution and 10\% of train samples are noises in the head classes
\item \textbf{\textit{VisA-step-K4}}: VisA dataset where its class distribution is imbalanced such that 7 out of 12 classes have only 4 samples for each class and 10\% of train samples are noises in the head classes
\item \textbf{\textit{VisA-step-K1}}: VisA dataset where its class distribution is imbalanced such that 7 out of 12 classes have only 1 samples for each class and 10\% of train samples are noises in the head classes
\end{itemize}

\section{Full Experiment Results}
\label{supp_sec:exp}

The full experiment results for the baseline comparison are given in Tabs. ~\ref{table:supp_img_mvtec_pareto}, ~\ref{table:supp_img_mvtec_k4}, ~\ref{table:supp_img_mvtec_k1}, ~\ref{table:supp_pix_mvtec_pareto}, ~\ref{table:supp_pix_mvtec_k4}, ~\ref{table:supp_pix_mvtec_k1},~\ref{table:supp_img_visa_pareto}, ~\ref{table:supp_img_visa_k4}, ~\ref{table:supp_img_visa_k1}, ~\ref{table:supp_pix_visa_pareto}, ~\ref{table:supp_pix_visa_k4}, and ~\ref{table:supp_pix_visa_k1}. We note that both WinCLIP and AnomalyCLIP are zero-shot models, hence they do not have performance variance dependant on the train sets (indicated without standard deviation in the tables).

\begin{table*}[t]
\caption{The metadata of MVTecAD-Pareto dataset, indicating the number of total samples in each class and the number of defect samples in each class in the train set.}
\label{table:supp_num_mvtec_pareto}
    \centering
    \renewcommand{\arraystretch}{1.2}
    {\fontsize{8}{9}\selectfont
    \resizebox{0.95\linewidth}{!}{
\begin{tabular}{ll|ccccccccccccccc}
    \toprule
        seed & ~ & bottle & cable & capsule & carpet & grid & hazelnut & leather & metal nut & pill & screw & tile & toothbrush & transistor & wood & zipper \\ \midrule
        101 & total & 5 & 5 & 5 & 180 & 297 & 5 & 31 & 22 & 46 & 10 & 69 & 14 & 110 & 5 & 7 \\
        ~ & anomaly & 0 & 0 & 0 & 20 & 33 & 0 & 2 & 2 & 3 & 0 & 4 & 0 & 10 & 0 & 0 \\ \midrule
        102 & total & 6 & 5 & 5 & 145 & 11 & 5 & 8 & 24 & 59 & 16 & 5 & 40 & 239 & 5 & 86 \\
        ~ & anomaly & 0 & 0 & 0 & 16 & 0 & 0 & 0 & 1 & 7 & 0 & 0 & 5 & 26 & 0 & 5 \\ \midrule
        103 & total & 231 & 11 & 8 & 5 & 145 & 5 & 24 & 40 & 84 & 6 & 16 & 5 & 5 & 5 & 58 \\
        ~ & anomaly & 22 & 0 & 0 & 0 & 18 & 0 & 1 & 6 & 5 & 0 & 0 & 0 & 0 & 0 & 7 \\ \midrule
        104 & total & 12 & 38 & 5 & 7 & 5 & 58 & 9 & 5 & 139 & 355 & 86 & 17 & 26 & 5 & 212 \\
        ~ & anomaly & 0 & 3 & 0 & 0 & 0 & 6 & 0 & 0 & 17 & 35 & 8 & 0 & 2 & 0 & 18 \\ \midrule
        105 & total & 15 & 47 & 5 & 308 & 8 & 22 & 5 & 11 & 5 & 190 & 5 & 6 & 32 & 79 & 122 \\
        ~ & anomaly & 0 & 2 & 0 & 28 & 0 & 1 & 0 & 0 & 0 & 20 & 0 & 0 & 1 & 10 & 16 \\ \bottomrule
    \end{tabular}
    }
    }
\end{table*}

\begin{table*}[t]
\caption{The metadata of MVTecAD-step-K4 dataset, indicating the number of total samples in each class and the number of defect samples in each class in the train set.}
\label{table:supp_num_mvtec_k4}
    \centering
    \renewcommand{\arraystretch}{1.2}
    {\fontsize{8}{9}\selectfont
    \resizebox{0.95\linewidth}{!}{
    \begin{tabular}{ll|ccccccccccccccc}
    \toprule
        seed & ~ & bottle & cable & capsule & carpet & grid & hazelnut & leather & metal nut & pill & screw & tile & toothbrush & transistor & wood & zipper \\ \midrule
        101 & total & 4 & 246 & 241 & 4 & 4 & 430 & 4 & 4 & 4 & 352 & 253 & 66 & 4 & 4 & 4 \\
        ~ & anomaly & 0 & 22 & 22 & 0 & 0 & 39 & 0 & 0 & 0 & 32 & 23 & 6 & 0 & 0 & 0 \\ \midrule
        102 & total & 4 & 4 & 4 & 4 & 4 & 4 & 4 & 242 & 294 & 352 & 4 & 66 & 4 & 272 & 264 \\
        ~ & anomaly & 0 & 0 & 0 & 0 & 0 & 0 & 0 & 22 & 27 & 32 & 0 & 6 & 0 & 25 & 24 \\ \midrule
        103 & total & 4 & 4 & 241 & 4 & 290 & 4 & 269 & 4 & 294 & 4 & 4 & 4 & 4 & 272 & 264 \\
        ~ & anomaly & 0 & 0 & 22 & 0 & 26 & 0 & 24 & 0 & 27 & 0 & 0 & 0 & 0 & 25 & 24 \\ \midrule
        104 & total & 4 & 4 & 4 & 308 & 4 & 430 & 269 & 4 & 4 & 352 & 253 & 4 & 4 & 4 & 264 \\
        ~ & anomaly & 0 & 0 & 0 & 28 & 0 & 39 & 24 & 0 & 0 & 32 & 23 & 0 & 0 & 0 & 24 \\ \midrule
        105 & total & 230 & 246 & 4 & 4 & 4 & 430 & 269 & 4 & 4 & 352 & 4 & 4 & 234 & 4 & 4 \\
        ~ & anomaly & 21 & 22 & 0 & 0 & 0 & 39 & 24 & 0 & 0 & 32 & 0 & 0 & 21 & 0 & 0 \\ \bottomrule
    \end{tabular}
    }
    }
\end{table*}

\begin{table*}[t]
\caption{The metadata of MVTecAD-step-K1 dataset, indicating the number of total samples in each class and the number of defect samples in each class in the train set.}
\label{table:supp_num_mvtec_k1}
    \centering
    \renewcommand{\arraystretch}{1.2}
    {\fontsize{8}{9}\selectfont
    \resizebox{0.95\linewidth}{!}{
    \begin{tabular}{ll|ccccccccccccccc}
    \toprule
        seed & ~ & bottle & cable & capsule & carpet & grid & hazelnut & leather & metal nut & pill & screw & tile & toothbrush & transistor & wood & zipper \\ \midrule
        101 & total & 1 & 1 & 1 & 308 & 1 & 430 & 269 & 1 & 1 & 352 & 253 & 1 & 1 & 1 & 264 \\
        ~ & anomaly & 0 & 0 & 0 & 28 & 0 & 39 & 24 & 0 & 0 & 32 & 23 & 0 & 0 & 0 & 24 \\ \midrule
        102 & total & 1 & 1 & 241 & 1 & 290 & 1 & 269 & 1 & 294 & 1 & 1 & 1 & 1 & 272 & 264 \\
        ~ & anomaly & 0 & 0 & 22 & 0 & 26 & 0 & 24 & 0 & 27 & 0 & 0 & 0 & 0 & 25 & 24 \\ \midrule
        103 & total & 1 & 1 & 241 & 1 & 1 & 1 & 269 & 1 & 1 & 352 & 1 & 66 & 234 & 272 & 1 \\
        ~ & anomaly & 0 & 0 & 22 & 0 & 0 & 0 & 24 & 0 & 0 & 32 & 0 & 6 & 21 & 25 & 0 \\ \midrule
        104 & total & 1 & 1 & 1 & 1 & 290 & 1 & 269 & 242 & 1 & 352 & 1 & 66 & 1 & 1 & 264 \\
        ~ & anomaly & 0 & 0 & 0 & 0 & 26 & 0 & 24 & 22 & 0 & 32 & 0 & 6 & 0 & 0 & 24 \\ \midrule
        105 & total & 1 & 246 & 241 & 1 & 290 & 430 & 1 & 1 & 294 & 352 & 1 & 1 & 1 & 1 & 1 \\
        ~ & anomaly & 0 & 22 & 22 & 0 & 26 & 39 & 0 & 0 & 27 & 32 & 0 & 0 & 0 & 0 & 0 \\ \bottomrule
    \end{tabular}
    }
    }
\end{table*}

\begin{table*}[t]
\caption{The metadata of VisA-Pareto dataset, indicating the number of total samples in each class and the number of defect samples in each class in the train set.}
\label{table:supp_num_visa_pareto}
    \centering
    \renewcommand{\arraystretch}{1.2}
    {\fontsize{8}{9}\selectfont
    \resizebox{0.95\linewidth}{!}{
    \begin{tabular}{ll|cccccccccccc}
    \toprule
        seed & ~ & candle & capsules & cashew & chewinggum & fryum & macaroni1 & macaroni2 & pcb1 & pcb2 & pcb3 & pcb4 & pipe fryum \\ \midrule
        101 & total & 290 & 13 & 476 & 9 & 19 & 943 & 29 & 105 & 173 & 68 & 43 & 7 \\
        ~ & anomaly & 14 & 0 & 26 & 0 & 0 & 43 & 1 & 5 & 10 & 4 & 1 & 0 \\ \midrule
        102 & total & 92 & 5 & 10 & 475 & 53 & 33 & 7 & 256 & 14 & 146 & 21 & 5 \\
        ~ & anomaly & 10 & 0 & 0 & 22 & 2 & 1 & 0 & 11 & 0 & 7 & 0 & 0 \\ \midrule
        103 & total & 5 & 9 & 86 & 21 & 479 & 253 & 5 & 145 & 7 & 33 & 14 & 52 \\
        ~ & anomaly & 0 & 0 & 4 & 0 & 29 & 10 & 0 & 7 & 0 & 1 & 0 & 2 \\ \midrule
        104 & total & 170 & 9 & 108 & 65 & 476 & 19 & 30 & 13 & 43 & 954 & 289 & 7 \\
        ~ & anomaly & 6 & 0 & 7 & 1 & 26 & 0 & 2 & 0 & 1 & 49 & 12 & 0 \\ \midrule
        105 & total & 29 & 101 & 43 & 19 & 177 & 297 & 942 & 13 & 9 & 7 & 67 & 471 \\
        ~ & anomaly & 1 & 1 & 1 & 0 & 14 & 21 & 42 & 0 & 0 & 0 & 3 & 21 \\ \bottomrule
    \end{tabular}
    }
    }
\end{table*}

\begin{table*}[t]
\caption{The metadata of VisA-step-K4 dataset, indicating the number of total samples in each class and the number of defect samples in each class in the train set.}
\label{table:supp_num_visa_k4}
    \centering
    \renewcommand{\arraystretch}{1.2}
    {\fontsize{8}{9}\selectfont
    \resizebox{0.95\linewidth}{!}{
    \begin{tabular}{ll|cccccccccccc}
    \toprule
        seed & ~ & candle & capsules & cashew & chewinggum & fryum & macaroni1 & macaroni2 & pcb1 & pcb2 & pcb3 & pcb4 & pipe fryum \\ \midrule
        101 & total & 4 & 4 & 4 & 4 & 4 & 945 & 4 & 4 & 946 & 950 & 949 & 472 \\
        ~ & anomaly & 0 & 0 & 0 & 0 & 0 & 45 & 0 & 0 & 45 & 45 & 45 & 22 \\ \midrule
        102 & total & 4 & 4 & 472 & 4 & 4 & 945 & 945 & 4 & 946 & 4 & 4 & 472 \\
        ~ & anomaly & 0 & 0 & 22 & 0 & 0 & 45 & 45 & 0 & 45 & 0 & 0 & 22 \\ \midrule
        103 & total & 4 & 4 & 4 & 4 & 4 & 4 & 4 & 949 & 946 & 950 & 949 & 472 \\
        ~ & anomaly & 0 & 0 & 0 & 0 & 0 & 0 & 0 & 45 & 45 & 45 & 45 & 22 \\ \midrule
        104 & total & 945 & 4 & 472 & 476 & 472 & 4 & 4 & 949 & 4 & 4 & 4 & 4 \\
        ~ & anomaly & 45 & 0 & 22 & 23 & 22 & 0 & 0 & 45 & 0 & 0 & 0 & 0 \\ \midrule
        105 & total & 945 & 4 & 472 & 476 & 4 & 945 & 4 & 949 & 4 & 4 & 4 & 4 \\
        ~ & anomaly & 45 & 0 & 22 & 23 & 0 & 45 & 0 & 45 & 0 & 0 & 0 & 0 \\ \bottomrule
    \end{tabular}
    }
    }
\end{table*}

\begin{table*}[t]
\caption{The metadata of VisA-step-K1 dataset, indicating the number of total samples in each class and the number of defect samples in each class in the train set.}
\label{table:supp_num_visa_k1}
    \centering
    \renewcommand{\arraystretch}{1.2}
    {\fontsize{8}{9}\selectfont
    \resizebox{0.95\linewidth}{!}{
    \begin{tabular}{ll|cccccccccccc}
    \toprule
        seed & ~ & candle & capsules & cashew & chewinggum & fryum & macaroni1 & macaroni2 & pcb1 & pcb2 & pcb3 & pcb4 & pipe fryum \\ \midrule
        101 & total & 1 & 1 & 1 & 1 & 1 & 945 & 1 & 1 & 946 & 950 & 949 & 472 \\
        ~ & anomaly & 0 & 0 & 0 & 0 & 0 & 45 & 0 & 0 & 45 & 45 & 45 & 22 \\ \midrule
        102 & total & 945 & 1 & 1 & 1 & 1 & 945 & 1 & 1 & 946 & 1 & 949 & 472 \\
        ~ & anomaly & 45 & 0 & 0 & 0 & 0 & 45 & 0 & 0 & 45 & 0 & 45 & 22 \\ \midrule
        103 & total & 1 & 569 & 472 & 1 & 472 & 945 & 1 & 1 & 946 & 1 & 1 & 1 \\
        ~ & anomaly & 0 & 27 & 22 & 0 & 22 & 45 & 0 & 0 & 45 & 0 & 0 & 0 \\ \midrule
        104 & total & 1 & 1 & 1 & 1 & 1 & 1 & 1 & 949 & 946 & 950 & 949 & 472 \\
        ~ & anomaly & 0 & 0 & 0 & 0 & 0 & 0 & 0 & 45 & 45 & 45 & 45 & 22 \\ \midrule
        105 & total & 1 & 569 & 472 & 1 & 472 & 1 & 1 & 949 & 1 & 950 & 1 & 1 \\
        ~ & anomaly & 0 & 27 & 22 & 0 & 22 & 0 & 0 & 45 & 0 & 45 & 0 & 0 \\ \bottomrule
    \end{tabular}
    }
    }
\end{table*}

\begin{table*}[t]
\caption{Anomaly detection on MVTecAD-Pareto with image-level AUROC. We report the mean over 5 random seeds for each measurement. The best performance is indicated with bold and the second-best with underline.}
\label{table:supp_img_mvtec_pareto}
    \centering
    \renewcommand{\arraystretch}{1.2}
    {\fontsize{8}{9}\selectfont
    \resizebox{\linewidth}{!}{
    \begin{tabular}{@{}l@{} | @{\hspace{2pt}}c@{\hspace{2pt}} @{\hspace{2pt}}c@{\hspace{2pt}} @{\hspace{2pt}}c@{\hspace{2pt}} @{\hspace{2pt}}c@{\hspace{2pt}} @{\hspace{2pt}}c@{\hspace{2pt}} @{\hspace{2pt}}c@{\hspace{2pt}} @{\hspace{2pt}}c@{\hspace{2pt}} @{\hspace{2pt}}c@{\hspace{2pt}} @{\hspace{2pt}}c@{\hspace{2pt}} @{\hspace{2pt}}c@{\hspace{2pt}}@{}}
    \toprule
        & PaDiM & SimpleNet & EfficientAD & HVQ & UniAD & WinCLIP & AnomalyCLIP & PatchCore & SoftPatch & \makecell{\ours\\(ours)} \\ \midrule
        bottle & 96.95 $\pm$ 2.89 & 79.89 $\pm$ 15.76 & 97.35 $\pm$ 3.28 & 92.65 $\pm$ 9.56 & 99.30 $\pm$ 1.04 & 98.60  & 89.30  & 92.19 $\pm$ 13.68 & 97.95 $\pm$ 4.14 & \textbf{99.76} $\pm$ 0.53 \\
        cable & 66.76 $\pm$ 4.43 & 65.19 $\pm$ 9.65 & 70.05 $\pm$ 17.25 & 70.28 $\pm$ 10.73 & 73.73 $\pm$ 12.36 & 85.00  & 69.80  & \textbf{94.81} $\pm$ 1.14 & 78.89 $\pm$ 19.04 & 94.54 $\pm$ 2.12 \\
        capsule & 67.33 $\pm$ 7.69 & 70.59 $\pm$ 7.45 & 69.23 $\pm$ 6.25 & 59.89 $\pm$ 2.36 & 64.25 $\pm$ 5.15 & 68.70  & 89.90 & 89.84 $\pm$ 3.78 & 58.78 $\pm$ 23.22 & \textbf{92.19} $\pm$ 1.67 \\
        carpet & 97.51 $\pm$ 4.93 & 78.74 $\pm$ 12.14 & 89.24 $\pm$ 10.07 & 98.48 $\pm$ 0.80 & 98.48 $\pm$ 2.41 & 99.30  & \textbf{100.00}  & 86.43 $\pm$ 12.97 & 99.11 $\pm$ 0.61 & 99.26 $\pm$ 0.29 \\
        grid & 64.31 $\pm$ 11.57 & 52.45 $\pm$ 2.66 & 77.61 $\pm$ 10.34 & 78.91 $\pm$ 7.71 & 87.54 $\pm$ 9.00 & \textbf{99.20}  & 97.00  & 64.04 $\pm$ 17.93 & 87.10 $\pm$ 9.44 & 91.33 $\pm$ 4.74 \\
        hazelnut & 77.87 $\pm$ 21.57 & 72.94 $\pm$ 10.23 & 95.14 $\pm$ 2.48 & 94.83 $\pm$ 3.12 & 93.66 $\pm$ 4.30 & 92.30  & 97.20 & 84.54 $\pm$ 25.95 & 94.26 $\pm$ 7.47 & \textbf{99.28} $\pm$ 0.83 \\
        leather & 99.91 $\pm$ 0.14 & 65.63 $\pm$ 15.84 & 90.34 $\pm$ 10.90 & 99.99 $\pm$ 0.02 & \textbf{100.00} $\pm$ 0.00 & \textbf{100.00}  & 99.80  & 99.35 $\pm$ 0.97 & 99.93 $\pm$ 0.15 & \textbf{100.00} $\pm$ 0.00 \\
\bottomrule
    \end{tabular}
    }
    }
\end{table*}

\begin{table*}[t]
\caption{Anomaly detection on MVTecAD-step-K4 with image-level AUROC. We report the mean over 5 random seeds for each measurement. The best performance is indicated with bold and the second-best with underline.}
\label{table:supp_img_mvtec_k4}
    \centering
    \renewcommand{\arraystretch}{1.2}
    {\fontsize{8}{9}\selectfont
    \resizebox{\linewidth}{!}{
    \begin{tabular}{@{}l@{} | @{\hspace{2pt}}c@{\hspace{2pt}} @{\hspace{2pt}}c@{\hspace{2pt}} @{\hspace{2pt}}c@{\hspace{2pt}} @{\hspace{2pt}}c@{\hspace{2pt}} @{\hspace{2pt}}c@{\hspace{2pt}} @{\hspace{2pt}}c@{\hspace{2pt}} @{\hspace{2pt}}c@{\hspace{2pt}} @{\hspace{2pt}}c@{\hspace{2pt}} @{\hspace{2pt}}c@{\hspace{2pt}} @{\hspace{2pt}}c@{\hspace{2pt}}@{}}
    \toprule
         & PaDiM & SimpleNet & EfficientAD & HVQ & UniAD & WinCLIP & AnomalyCLIP & PatchCore & SoftPatch & \makecell{\ours\\(ours)} \\ \midrule
        bottle & 94.30 $\pm$ 5.03 & 74.36 $\pm$ 18.09 & 90.45 $\pm$ 10.43 & 76.78 $\pm$ 7.92 & 96.98 $\pm$ 3.02 & 98.60  & 89.30  & 95.49 $\pm$ 9.50 & 65.92 $\pm$ 27.24 & \textbf{99.75} $\pm$ 0.26 \\ 
        cable & 63.05 $\pm$ 5.20 & 57.45 $\pm$ 7.96 & 75.18 $\pm$ 12.58 & 67.10 $\pm$ 21.09 & 71.47 $\pm$ 19.40 & 85.00  & 69.80  & 85.35 $\pm$ 9.56 & 72.85 $\pm$ 24.09 & \textbf{96.18} $\pm$ 2.87 \\ 
        capsule & 67.41 $\pm$ 16.32 & 62.25 $\pm$ 14.52 & 67.40 $\pm$ 27.08 & 82.37 $\pm$ 17.80 & 69.63 $\pm$ 16.31 & 68.70  & 89.90  & 89.18 $\pm$ 3.35 & 64.54 $\pm$ 28.56 & \textbf{93.59} $\pm$ 2.67 \\ 
        carpet & 99.62 $\pm$ 0.39 & 74.38 $\pm$ 8.92 & 89.07 $\pm$ 7.38 & 98.17 $\pm$ 0.83 & 99.63 $\pm$ 0.47 & 99.30  & \textbf{100.00}  & 92.32 $\pm$ 13.43 & 96.90 $\pm$ 1.84 & 99.24 $\pm$ 0.48 \\ 
        grid & 71.70 $\pm$ 3.08 & 53.85 $\pm$ 14.11 & 66.57 $\pm$ 20.92 & 84.85 $\pm$ 3.63 & 89.49 $\pm$ 4.82 & \textbf{99.20}  & 97.00  & 68.60 $\pm$ 8.93 & 74.34 $\pm$ 14.00 & 93.38 $\pm$ 3.63 \\ 
        hazelnut & 78.01 $\pm$ 19.31 & 71.68 $\pm$ 5.07 & 92.24 $\pm$ 7.56 & 84.18 $\pm$ 8.84 & 92.84 $\pm$ 7.23 & 92.30  & \textbf{97.20}  & 63.86 $\pm$ 27.00 & 92.86 $\pm$ 9.85 & 96.67 $\pm$ 4.03 \\ 
        leather & 99.71 $\pm$ 0.65 & 55.22 $\pm$ 16.24 & 93.43 $\pm$ 10.23 & 98.30 $\pm$ 3.75 & 100.00 $\pm$ 0.00 & 100.00  & 99.80  & 85.99 $\pm$ 13.06 & 97.00 $\pm$ 6.70 & \textbf{100.00} $\pm$ 0.00 \\
\bottomrule
    \end{tabular}
    }
    }
\end{table*}

\begin{table*}[t]
\caption{Anomaly detection on MVTecAD-step-K1 with image-level AUROC. We report the mean over 5 random seeds for each measurement. The best performance is indicated with bold and the second-best with underline.}
\label{table:supp_img_mvtec_k1}
    \centering
    \renewcommand{\arraystretch}{1.2}
    {\fontsize{8}{9}\selectfont
    \resizebox{\linewidth}{!}{
    \begin{tabular}{@{}l@{} | @{\hspace{2pt}}c@{\hspace{2pt}} @{\hspace{2pt}}c@{\hspace{2pt}} @{\hspace{2pt}}c@{\hspace{2pt}} @{\hspace{2pt}}c@{\hspace{2pt}} @{\hspace{2pt}}c@{\hspace{2pt}} @{\hspace{2pt}}c@{\hspace{2pt}} @{\hspace{2pt}}c@{\hspace{2pt}} @{\hspace{2pt}}c@{\hspace{2pt}} @{\hspace{2pt}}c@{\hspace{2pt}} @{\hspace{2pt}}c@{\hspace{2pt}}@{}}
    \toprule
         & PaDiM & SimpleNet & EfficientAD & HVQ & UniAD & WinCLIP & AnomalyCLIP & PatchCore & SoftPatch & \makecell{\ours\\(ours)} \\ \midrule
        bottle & 72.68 $\pm$ 15.75 & 60.40 $\pm$ 18.58 & 57.32 $\pm$ 17.54 & 73.71 $\pm$ 13.23 & 90.74 $\pm$ 3.63 & 98.60  & 89.30  & \textbf{99.65} $\pm$ 0.12 & 36.13 $\pm$ 3.18 & 99.25 $\pm$ 0.52 \\
        cable & 57.79 $\pm$ 9.61 & 55.09 $\pm$ 10.91 & 55.21 $\pm$ 19.54 & 58.63 $\pm$ 12.98 & 71.25 $\pm$ 18.42 & 85.00  & 69.80  & 76.14 $\pm$ 7.38 & 61.97 $\pm$ 20.97 & \textbf{89.89} $\pm$ 4.87 \\
        capsule & 78.92 $\pm$ 17.70 & 66.29 $\pm$ 13.88 & 65.17 $\pm$ 24.08 & 72.48 $\pm$ 19.26 & 76.86 $\pm$ 13.02 & 68.70  & 89.90  & 82.24 $\pm$ 4.85 & 72.55 $\pm$ 32.06 & \textbf{90.39} $\pm$ 7.63 \\
        carpet & 98.53 $\pm$ 0.78 & 57.46 $\pm$ 16.18 & 55.02 $\pm$ 17.19 & 97.88 $\pm$ 1.02 & 99.84 $\pm$ 0.10 & 99.30  & \textbf{100.00}  & 92.01 $\pm$ 13.37 & 92.58 $\pm$ 6.85 & 99.13 $\pm$ 0.29 \\
        grid & 67.84 $\pm$ 5.25 & 53.55 $\pm$ 12.45 & 73.28 $\pm$ 24.70 & 83.01 $\pm$ 5.46 & 94.02 $\pm$ 4.83 & \textbf{99.20}  & 97.00  & 59.62 $\pm$ 7.35 & 88.69 $\pm$ 11.25 & 91.66 $\pm$ 4.84 \\
        hazelnut & 77.88 $\pm$ 18.88 & 63.34 $\pm$ 6.59 & 69.95 $\pm$ 20.92 & 87.35 $\pm$ 7.10 & 90.43 $\pm$ 6.63 & 92.30  & \textbf{97.20}  & 61.91 $\pm$ 16.15 & 84.68 $\pm$ 15.04 & 94.81 $\pm$ 4.08 \\
        leather & 99.14 $\pm$ 1.78 & 54.68 $\pm$ 9.26 & 73.16 $\pm$ 23.74 & 99.99 $\pm$ 0.03 & 99.99 $\pm$ 0.02 & \textbf{100.00}  & 99.80  & 79.23 $\pm$ 11.61 & 89.29 $\pm$ 23.94 & \textbf{100.00} $\pm$ 0.00 \\
\bottomrule
    \end{tabular}
    }
    }
\end{table*}

\begin{table*}[t]
\caption{Anomaly segmentation on MVTecAD-Pareto with pixel-level AUROC. We report the mean over 5 random seeds for each measurement. The best performance is indicated with bold and the second-best with underline.}
\label{table:supp_pix_mvtec_pareto}
    \centering
    \renewcommand{\arraystretch}{1.2}
    {\fontsize{8}{9}\selectfont
    \resizebox{\linewidth}{!}{
    \begin{tabular}{@{}l@{} | @{\hspace{2pt}}c@{\hspace{2pt}} @{\hspace{2pt}}c@{\hspace{2pt}} @{\hspace{2pt}}c@{\hspace{2pt}} @{\hspace{2pt}}c@{\hspace{2pt}} @{\hspace{2pt}}c@{\hspace{2pt}} @{\hspace{2pt}}c@{\hspace{2pt}} @{\hspace{2pt}}c@{\hspace{2pt}} @{\hspace{2pt}}c@{\hspace{2pt}} @{\hspace{2pt}}c@{\hspace{2pt}} @{\hspace{2pt}}c@{\hspace{2pt}}@{}}
    \toprule
         & PaDiM & SimpleNet & EfficientAD & HVQ & UniAD & WinCLIP & AnomalyCLIP & PatchCore & SoftPatch & \makecell{\ours\\(ours)} \\ \midrule
        bottle & 95.65 $\pm$ 1.24 & 75.95 $\pm$ 11.28 & 95.66 $\pm$ 2.28 & 96.24 $\pm$ 2.71 & \textbf{97.71} $\pm$ 0.40 & 85.70  & 90.40  & 91.32 $\pm$ 12.81 & 97.66 $\pm$ 1.52 & 97.31 $\pm$ 2.64 \\
        cable & 82.43 $\pm$ 6.33 & 66.77 $\pm$ 7.52 & 87.41 $\pm$ 11.64 & 89.50 $\pm$ 3.77 & \textbf{92.45} $\pm$ 2.08 & 61.30  & 78.90  & 90.68 $\pm$ 2.55 & 87.09 $\pm$ 9.27 & 91.43 $\pm$ 2.33 \\
        capsule & 97.14 $\pm$ 0.58 & 82.35 $\pm$ 7.93 & 94.96 $\pm$ 2.95 & 97.02 $\pm$ 0.22 & 96.82 $\pm$ 0.74 & 87.00  & 95.80  & \textbf{98.51} $\pm$ 0.10 & 94.72 $\pm$ 3.52 & 98.37 $\pm$ 0.09 \\
        carpet & 96.40 $\pm$ 3.98 & 71.85 $\pm$ 16.78 & 88.95 $\pm$ 9.43 & 98.51 $\pm$ 0.39 & 98.43 $\pm$ 0.23 & 90.90  & 98.80  & 85.87 $\pm$ 13.89 & 98.89 $\pm$ 0.18 & \textbf{98.48} $\pm$ 0.85 \\
        grid & 67.11 $\pm$ 21.48 & 35.73 $\pm$ 17.39 & 74.45 $\pm$ 11.12 & 92.70 $\pm$ 4.10 & 92.33 $\pm$ 3.01 & 79.40  & \textbf{97.30}  & 64.41 $\pm$ 15.35 & 74.97 $\pm$ 27.60 & 91.09 $\pm$ 4.28 \\
        hazelnut & 94.78 $\pm$ 0.88 & 86.32 $\pm$ 5.06 & \textbf{97.15} $\pm$ 0.98 & 96.36 $\pm$ 0.98 & 95.63 $\pm$ 1.26 & 95.70  & 97.10  & 83.29 $\pm$ 24.37 & 95.44 $\pm$ 1.56 & 96.37 $\pm$ 4.75 \\
        leather & 99.08 $\pm$ 0.15 & 68.17 $\pm$ 34.40 & 96.07 $\pm$ 3.19 & 98.89 $\pm$ 0.13 & 99.10 $\pm$ 0.08 & 95.50  & 98.60  & 99.19 $\pm$ 0.15 & 99.27 $\pm$ 0.04 & \textbf{99.29} $\pm$ 00.03 \\
\bottomrule
    \end{tabular}
    }
    }
\end{table*}

\begin{table*}[t]
\caption{Anomaly segmentation on MVTecAD-step-K4 with pixel-level AUROC. We report the mean over 5 random seeds for each measurement. The best performance is indicated with bold and the second-best with underline.}
\label{table:supp_pix_mvtec_k4}
    \centering
    \renewcommand{\arraystretch}{1.2}
    {\fontsize{8}{9}\selectfont
    \resizebox{\linewidth}{!}{
    \begin{tabular}{@{}l@{} | @{\hspace{2pt}}c@{\hspace{2pt}} @{\hspace{2pt}}c@{\hspace{2pt}} @{\hspace{2pt}}c@{\hspace{2pt}} @{\hspace{2pt}}c@{\hspace{2pt}} @{\hspace{2pt}}c@{\hspace{2pt}} @{\hspace{2pt}}c@{\hspace{2pt}} @{\hspace{2pt}}c@{\hspace{2pt}} @{\hspace{2pt}}c@{\hspace{2pt}} @{\hspace{2pt}}c@{\hspace{2pt}} @{\hspace{2pt}}c@{\hspace{2pt}}@{}}
    \toprule
         & PaDiM & SimpleNet & EfficientAD & HVQ & UniAD & WinCLIP & AnomalyCLIP & PatchCore & SoftPatch & \makecell{\ours\\(ours)} \\ \midrule
        bottle & 91.44 $\pm$ 5.59 & 70.98 $\pm$ 6.76 & 91.71 $\pm$ 4.77 & 91.42 $\pm$ 4.94 & 96.66 $\pm$ 1.00 & 85.70  & 90.40  & 93.79 $\pm$ 10.21 & 80.67 $\pm$ 16.93 & \textbf{98.44} $\pm$ 0.06 \\
        cable & 76.19 $\pm$ 11.95 & 69.18 $\pm$ 8.88 & 90.60 $\pm$ 8.38 & 83.80 $\pm$ 9.04 & 90.16 $\pm$ 5.65 & 61.30  & 78.90  & 83.20 $\pm$ 10.16 & 78.37 $\pm$ 17.42 & \textbf{92.70} $\pm$ 3.49 \\
        capsule & 96.29 $\pm$ 1.79 & 78.90 $\pm$ 13.70 & 90.40 $\pm$ 9.88 & 97.65 $\pm$ 2.17 & 96.55 $\pm$ 2.04 & 87.00  & 95.80  & 92.31 $\pm$ 8.51 & 93.48 $\pm$ 4.96 & \textbf{98.66} $\pm$ 0.15 \\
        carpet & 97.26 $\pm$ 1.57 & 63.79 $\pm$ 34.82 & 93.59 $\pm$ 1.19 & 96.59 $\pm$ 1.85 & 98.49 $\pm$ 0.16 & 90.90  & 98.80  & 93.03 $\pm$ 13.35 & 98.70 $\pm$ 0.35 & \textbf{99.09} $\pm$ 0.11 \\
        grid & 53.43 $\pm$ 26.71 & 24.65 $\pm$ 5.41 & 68.62 $\pm$ 18.87 & 94.05 $\pm$ 7.42 & 91.61 $\pm$ 3.53 & 79.40  & \textbf{97.30}  & 68.15 $\pm$ 2.89 & 52.00 $\pm$ 28.50 & 90.66 $\pm$ 5.13 \\
        hazelnut & 95.35 $\pm$ 1.51 & 81.11 $\pm$ 10.20 & 96.28 $\pm$ 0.83 & 93.45 $\pm$ 3.30 & 95.86 $\pm$ 2.80 & 95.70  & \textbf{97.10}  & 66.24 $\pm$ 29.23 & 92.85 $\pm$ 2.00 & 85.22 $\pm$ 15.52 \\
        leather & 99.03 $\pm$ 0.26 & 62.90 $\pm$ 31.26 & 95.48 $\pm$ 4.09 & 98.06 $\pm$ 2.32 & 99.04 $\pm$ 0.17 & 95.50  & 98.60  & 82.26 $\pm$ 17.30 & 98.41 $\pm$ 2.01 & \textbf{99.30} $\pm$ 0.01 \\
\bottomrule
    \end{tabular}
    }
    }
\end{table*}

\begin{table*}[t]
\caption{Anomaly segmentation on MVTecAD-step-K1 with pixel-level AUROC. We report the mean over 5 random seeds for each measurement. The best performance is indicated with bold and the second-best with underline.}
\label{table:supp_pix_mvtec_k1}
    \centering
    \renewcommand{\arraystretch}{1.2}
    {\fontsize{8}{9}\selectfont
    \resizebox{\linewidth}{!}{
    \begin{tabular}{@{}l@{} | @{\hspace{2pt}}c@{\hspace{2pt}} @{\hspace{2pt}}c@{\hspace{2pt}} @{\hspace{2pt}}c@{\hspace{2pt}} @{\hspace{2pt}}c@{\hspace{2pt}} @{\hspace{2pt}}c@{\hspace{2pt}} @{\hspace{2pt}}c@{\hspace{2pt}} @{\hspace{2pt}}c@{\hspace{2pt}} @{\hspace{2pt}}c@{\hspace{2pt}} @{\hspace{2pt}}c@{\hspace{2pt}} @{\hspace{2pt}}c@{\hspace{2pt}}@{}}
    \toprule
         & PaDiM & SimpleNet & EfficientAD & HVQ & UniAD & WinCLIP & AnomalyCLIP & PatchCore & SoftPatch & \makecell{\ours\\(ours)} \\ \midrule
        bottle & 76.52 $\pm$ 11.38 & 65.85 $\pm$ 16.14 & 60.97 $\pm$ 16.90 & 85.23 $\pm$ 14.39 & 94.31 $\pm$ 0.57 & 85.70  & 90.40  & 98.13 $\pm$ 0.08 & 50.48 $\pm$ 7.86 & \textbf{98.36} $\pm$ 0.04 \\
        cable & 71.28 $\pm$ 16.25 & 65.36 $\pm$ 12.90 & 75.16 $\pm$ 13.19 & 78.51 $\pm$ 7.99 & \textbf{85.37} $\pm$ 7.80 & 61.30  & 78.90  & 76.87 $\pm$ 9.24 & 61.93 $\pm$ 19.83 & 83.29 $\pm$ 1.47 \\
        capsule & 97.17 $\pm$ 2.50 & 83.48 $\pm$ 10.86 & 84.41 $\pm$ 15.09 & 97.14 $\pm$ 2.20 & 97.10 $\pm$ 1.86 & 87.00  & 95.80  & 87.59 $\pm$ 9.90 & 94.80 $\pm$ 5.59 & \textbf{98.28} $\pm$ 0.77 \\
        carpet & 95.21 $\pm$ 3.11 & 41.13 $\pm$ 37.80 & 68.31 $\pm$ 19.53 & 97.68 $\pm$ 0.59 & 98.43 $\pm$ 0.06 & 90.90  & 98.80  & 93.04 $\pm$ 13.33 & 91.00 $\pm$ 11.06 & \textbf{99.04} $\pm$ 0.18 \\
        grid & 71.65 $\pm$ 27.37 & 31.36 $\pm$ 16.26 & 75.59 $\pm$ 18.92 & 94.20 $\pm$ 5.63 & 92.20 $\pm$ 4.56 & 79.40  & \textbf{97.30}  & 62.25 $\pm$ 6.55 & 68.87 $\pm$ 37.97 & 88.13 $\pm$ 9.06 \\
        hazelnut & 94.05 $\pm$ 1.74 & 76.90 $\pm$ 4.35 & 88.40 $\pm$ 9.31 & 94.66 $\pm$ 1.97 & 94.69 $\pm$ 2.81 & 95.70  & \textbf{97.10}  & 69.53 $\pm$ 32.77 & 90.52 $\pm$ 2.27 & 86.49 $\pm$ 15.74 \\
        leather & 98.66 $\pm$ 1.03 & 60.45 $\pm$ 27.40 & 80.45 $\pm$ 20.05 & 99.12 $\pm$ 0.09 & 99.06 $\pm$ 0.18 & 95.50  & 98.60  & 78.87 $\pm$ 13.04 & 94.94 $\pm$ 9.75 & \textbf{99.31} $\pm$ 0.01 \\
\bottomrule
    \end{tabular}
    }
    }
\end{table*}

\begin{table*}[t]
\caption{Anomaly detection on VisA-Pareto with image-level AUROC. We report the mean over 5 random seeds for each measurement. The best performance is indicated with bold and the second-best with underline.}
\label{table:supp_img_visa_pareto}
    \centering
    \renewcommand{\arraystretch}{1.2}
    {\fontsize{8}{10}\selectfont
    \resizebox{\linewidth}{!}{
    \begin{tabular}{@{}l@{} | @{\hspace{2pt}}c@{\hspace{2pt}} @{\hspace{2pt}}c@{\hspace{2pt}} @{\hspace{2pt}}c@{\hspace{2pt}} @{\hspace{2pt}}c@{\hspace{2pt}} @{\hspace{2pt}}c@{\hspace{2pt}} @{\hspace{2pt}}c@{\hspace{2pt}} @{\hspace{2pt}}c@{\hspace{2pt}} @{\hspace{2pt}}c@{\hspace{2pt}} @{\hspace{2pt}}c@{\hspace{2pt}} @{\hspace{2pt}}c@{\hspace{2pt}}@{}}
    \toprule
         & PaDiM & SimpleNet & EfficientAD & HVQ & UniAD & WinCLIP & AnomalyCLIP & PatchCore & SoftPatch & \makecell{\ours\\(ours)} \\ \midrule
        candle & 86.88 $\pm$ 7.30 & 80.80 $\pm$ 7.81 & 90.25 $\pm$ 6.25 & 91.63 $\pm$ 3.55 & 80.41 $\pm$ 12.16 & 94.90  & 79.30  & 92.12 $\pm$ 3.46 & 84.04 $\pm$ 22.34 & \textbf{95.95} $\pm$ 1.18 \\
        capsules & 57.52 $\pm$ 2.04 & 68.07 $\pm$ 5.44 & 66.71 $\pm$ 6.85 & 63.95 $\pm$ 3.76 & 55.32 $\pm$ 1.55 & 79.40  & 81.50  & 66.08 $\pm$ 6.72 & 62.24 $\pm$ 5.23 & \textbf{69.91} $\pm$ 5.36 \\
        cashew & 85.60 $\pm$ 3.49 & 84.66 $\pm$ 5.94 & 91.66 $\pm$ 5.64 & 87.09 $\pm$ 7.38 & 84.99 $\pm$ 9.83 & 91.20  & 76.30  & 89.59 $\pm$ 9.87 & 95.95 $\pm$ 0.41 & \textbf{96.18} $\pm$ 0.63 \\
        chewinggum & 88.61 $\pm$ 5.90 & 91.05 $\pm$ 8.40 & 96.94 $\pm$ 3.87 & 95.58 $\pm$ 2.28 & 96.06 $\pm$ 1.04 & 95.50  & 97.40  & 94.27 $\pm$ 8.69 & \textbf{98.61} $\pm$ 0.52 & 97.84 $\pm$ 1.04 \\
        fryum & 80.02 $\pm$ 6.04 & 77.24 $\pm$ 6.10 & 89.89 $\pm$ 3.25 & 84.05 $\pm$ 2.88 & 76.49 $\pm$ 3.20 & 73.60  & 93.00  & 83.30 $\pm$ 8.74 & 93.31 $\pm$ 1.45 & \textbf{93.96} $\pm$ 0.70 \\
        macaroni1 & 78.52 $\pm$ 7.18 & 77.59 $\pm$ 4.87 & 80.72 $\pm$ 9.84 & 65.83 $\pm$ 10.11 & 70.47 $\pm$ 7.12 & 79.00  & 87.20  & 81.91 $\pm$ 10.80 & 89.99 $\pm$ 7.90 & \textbf{92.11} $\pm$ 4.16 \\
        macaroni2 & 61.87 $\pm$ 6.68 & 52.30 $\pm$ 3.42 & 60.27 $\pm$ 6.57 & 62.94 $\pm$ 8.93 & 58.14 $\pm$ 6.88 & 67.10  & \textbf{73.40}  & 64.89 $\pm$ 11.32 & 56.65 $\pm$ 10.27 & 71.48 $\pm$ 2.94 \\
\bottomrule
    \end{tabular}
    }
    }
\end{table*}

\begin{table*}[t]
\caption{Anomaly detection on VisA-step-K4 with image-level AUROC. We report the mean over 5 random seeds for each measurement. The best performance is indicated with bold and the second-best with underline.}
\label{table:supp_img_visa_k4}
    \centering
    \renewcommand{\arraystretch}{1.2}
    {\fontsize{8}{10}\selectfont
    \resizebox{\linewidth}{!}{
    \begin{tabular}{@{}l@{} | @{\hspace{2pt}}c@{\hspace{2pt}} @{\hspace{2pt}}c@{\hspace{2pt}} @{\hspace{2pt}}c@{\hspace{2pt}} @{\hspace{2pt}}c@{\hspace{2pt}} @{\hspace{2pt}}c@{\hspace{2pt}} @{\hspace{2pt}}c@{\hspace{2pt}} @{\hspace{2pt}}c@{\hspace{2pt}} @{\hspace{2pt}}c@{\hspace{2pt}} @{\hspace{2pt}}c@{\hspace{2pt}} @{\hspace{2pt}}c@{\hspace{2pt}}@{}}
    \toprule
         & PaDiM & SimpleNet & EfficientAD & HVQ & UniAD & WinCLIP & AnomalyCLIP & PatchCore & SoftPatch & \makecell{\ours\\(ours)} \\ \midrule
        candle & 70.21 $\pm$ 21.15 & 72.35 $\pm$ 10.28 & 67.81 $\pm$ 26.77 & 86.47 $\pm$ 9.89 & 66.55 $\pm$ 18.77 & \textbf{94.90}  & 79.30  & 86.94 $\pm$ 8.04 & 66.09 $\pm$ 30.70 & 94.30 $\pm$ 2.66 \\
        capsules & 56.49 $\pm$ 1.12 & 68.11 $\pm$ 6.70 & 62.70 $\pm$ 6.64 & 59.55 $\pm$ 2.97 & 56.11 $\pm$ 2.86 & \textbf{79.40}  & 81.50  & 59.88 $\pm$ 2.99 & 58.50 $\pm$ 4.02 & 70.06 $\pm$ 2.45 \\
        cashew & 79.82 $\pm$ 15.68 & 77.46 $\pm$ 3.32 & 91.98 $\pm$ 6.17 & 82.54 $\pm$ 11.97 & 77.47 $\pm$ 20.14 & 91.20  & 76.30  & 83.38 $\pm$ 9.76 & 86.08 $\pm$ 15.53 & \textbf{95.42} $\pm$ 1.66 \\
        chewinggum & 75.35 $\pm$ 21.02 & 79.44 $\pm$ 3.80 & 85.94 $\pm$ 18.73 & 93.06 $\pm$ 4.91 & 89.45 $\pm$ 6.66 & 95.50  & 97.40  & 89.20 $\pm$ 11.90 & 93.92 $\pm$ 5.16 & \textbf{98.62} $\pm$ 0.64 \\
        fryum & 65.18 $\pm$ 12.47 & 75.26 $\pm$ 5.35 & 75.78 $\pm$ 12.23 & 72.47 $\pm$ 7.91 & 64.78 $\pm$ 9.52 & \textbf{73.60}  & 93.00  & 86.46 $\pm$ 6.80 & 50.93 $\pm$ 24.47 & 92.35 $\pm$ 1.82 \\
        macaroni1 & 76.63 $\pm$ 10.17 & 69.72 $\pm$ 10.58 & 74.02 $\pm$ 24.22 & 70.72 $\pm$ 10.86 & 68.49 $\pm$ 10.40 & 79.00  & 87.20  & 70.15 $\pm$ 8.37 & 78.08 $\pm$ 24.37 & \textbf{88.46} $\pm$ 9.69 \\
        macaroni2 & 59.73 $\pm$ 4.53 & 50.93 $\pm$ 5.62 & 63.14 $\pm$ 8.68 & 60.06 $\pm$ 7.47 & 55.81 $\pm$ 4.23 & 67.10  & \textbf{73.40}  & 60.19 $\pm$ 10.80 & 53.01 $\pm$ 10.33 & 69.59 $\pm$ 3.02 \\
\bottomrule
    \end{tabular}
    }
    }
\end{table*}

\begin{table*}[t]
\caption{Anomaly detection on VisA-step-K1 with image-level AUROC. We report the mean over 5 random seeds for each measurement. The best performance is indicated with bold and the second-best with underline.}
\label{table:supp_img_visa_k1}
    \centering
    \renewcommand{\arraystretch}{1.2}
    {\fontsize{8}{10}\selectfont
    \resizebox{\linewidth}{!}{
    \begin{tabular}{@{}l@{} | @{\hspace{2pt}}c@{\hspace{2pt}} @{\hspace{2pt}}c@{\hspace{2pt}} @{\hspace{2pt}}c@{\hspace{2pt}} @{\hspace{2pt}}c@{\hspace{2pt}} @{\hspace{2pt}}c@{\hspace{2pt}} @{\hspace{2pt}}c@{\hspace{2pt}} @{\hspace{2pt}}c@{\hspace{2pt}} @{\hspace{2pt}}c@{\hspace{2pt}} @{\hspace{2pt}}c@{\hspace{2pt}} @{\hspace{2pt}}c@{\hspace{2pt}}@{}}
    \toprule
         & PaDiM & SimpleNet & EfficientAD & HVQ & UniAD & WinCLIP & AnomalyCLIP & PatchCore & SoftPatch & \makecell{\ours\\(ours)} \\ \midrule
        candle & 51.53 $\pm$ 24.34 & 66.92 $\pm$ 19.26 & 54.68 $\pm$ 28.57 & 71.00 $\pm$ 15.55 & 74.89 $\pm$ 6.83 & \textbf{94.90}  & 79.30  & 87.41 $\pm$ 7.11 & 47.01 $\pm$ 28.30 & 92.33 $\pm$ 3.58 \\
        capsules & 55.24 $\pm$ 2.23 & 58.46 $\pm$ 4.42 & 64.51 $\pm$ 14.93 & 60.77 $\pm$ 6.42 & 53.56 $\pm$ 0.83 & 79.40  & \textbf{81.50}  & 54.37 $\pm$ 4.14 & 58.59 $\pm$ 5.55 & 64.33 $\pm$ 4.21 \\
        cashew & 71.51 $\pm$ 17.61 & 69.40 $\pm$ 12.10 & 83.28 $\pm$ 13.51 & 71.55 $\pm$ 18.13 & 71.12 $\pm$ 18.80 & 91.20  & 76.30  & 88.78 $\pm$ 4.47 & 75.20 $\pm$ 19.42 & \textbf{93.89} $\pm$ 3.03 \\
        chewinggum & 56.00 $\pm$ 10.76 & 65.94 $\pm$ 9.60 & 55.56 $\pm$ 3.73 & 67.93 $\pm$ 7.83 & 62.56 $\pm$ 9.82 & 95.50  & 97.40  & 97.30 $\pm$ 0.91 & 75.18 $\pm$ 15.66 & \textbf{97.97} $\pm$ 0.34 \\
        fryum & 67.91 $\pm$ 16.31 & 72.33 $\pm$ 8.71 & 73.71 $\pm$ 20.03 & 71.31 $\pm$ 13.56 & 64.88 $\pm$ 12.91 & 73.60  & \textbf{93.00}  & 82.99 $\pm$ 1.80 & 64.21 $\pm$ 28.02 & 91.84 $\pm$ 2.50 \\
        macaroni1 & 74.01 $\pm$ 12.30 & 63.53 $\pm$ 15.04 & 78.65 $\pm$ 17.86 & 66.28 $\pm$ 15.48 & 67.13 $\pm$ 12.11 & 79.00  & \textbf{87.20}  & 64.53 $\pm$ 6.42 & 76.81 $\pm$ 25.38 & 84.55 $\pm$ 14.14 \\
        macaroni2 & 58.82 $\pm$ 5.20 & 57.12 $\pm$ 4.10 & 45.77 $\pm$ 10.89 & 54.65 $\pm$ 3.86 & 59.91 $\pm$ 7.73 & 67.10  & \textbf{73.40}  & 58.12 $\pm$ 3.22 & 40.87 $\pm$ 5.62 & 66.36 $\pm$ 3.16 \\
\bottomrule
    \end{tabular}
    }
    }
\end{table*}

\begin{table*}[t]
\caption{Anomaly segmentation on VisA-Pareto with pixel-level AUROC. We report the mean over 5 random seeds for each measurement. The best performance is indicated with bold and the second-best with underline.}
\label{table:supp_pix_visa_pareto}
    \centering
    \renewcommand{\arraystretch}{1.2}
    {\fontsize{8}{10}\selectfont
    \resizebox{\linewidth}{!}{
    \begin{tabular}{@{}l@{} | @{\hspace{2pt}}c@{\hspace{2pt}} @{\hspace{2pt}}c@{\hspace{2pt}} @{\hspace{2pt}}c@{\hspace{2pt}} @{\hspace{2pt}}c@{\hspace{2pt}} @{\hspace{2pt}}c@{\hspace{2pt}} @{\hspace{2pt}}c@{\hspace{2pt}} @{\hspace{2pt}}c@{\hspace{2pt}} @{\hspace{2pt}}c@{\hspace{2pt}} @{\hspace{2pt}}c@{\hspace{2pt}} @{\hspace{2pt}}c@{\hspace{2pt}}@{}}
    \toprule
         & PaDiM & SimpleNet & EfficientAD & HVQ & UniAD & WinCLIP & AnomalyCLIP & PatchCore & SoftPatch & \makecell{\ours\\(ours)} \\ \midrule
        candle & 97.49 $\pm$ 1.24 & 92.06 $\pm$ 3.38 & 95.98 $\pm$ 0.88 & 98.63 $\pm$ 0.47 & 97.77 $\pm$ 1.38 & 87.00  & \textbf{98.80}  & 92.98 $\pm$ 6.06 & 93.92 $\pm$ 10.70 & 98.75 $\pm$ 0.63 \\
        capsules & 78.32 $\pm$ 5.30 & 91.73 $\pm$ 2.28 & 95.63 $\pm$ 2.66 & 91.77 $\pm$ 2.96 & 83.58 $\pm$ 1.76 & 80.00  & 95.00  & 95.92 $\pm$ 1.03 & 88.76 $\pm$ 10.75 & \textbf{97.68} $\pm$ 0.49 \\
        cashew & 95.78 $\pm$ 1.37 & 95.69 $\pm$ 1.05 & 97.48 $\pm$ 1.38 & \textbf{98.74} $\pm$ 0.38 & 98.50 $\pm$ 0.53 & 84.80  & 93.80  & 90.20 $\pm$ 11.64 & 98.02 $\pm$ 0.25 & 98.39 $\pm$ 0.22 \\
        chewinggum & 96.86 $\pm$ 1.00 & 94.28 $\pm$ 2.57 & 98.64 $\pm$ 0.50 & 98.54 $\pm$ 0.33 & 98.75 $\pm$ 0.10 & 95.40  & \textbf{99.30}  & 94.27 $\pm$ 7.88 & 98.05 $\pm$ 0.21 & 99.06 $\pm$ 0.52 \\
        fryum & 92.73 $\pm$ 1.60 & 85.64 $\pm$ 1.58 & 92.84 $\pm$ 2.57 & \textbf{97.16} $\pm$ 0.30 & 96.45 $\pm$ 1.01 & 87.70  & 94.60  & 82.65 $\pm$ 8.10 & 91.37 $\pm$ 0.56 & 91.61 $\pm$ 0.54 \\
        macaroni1 & 97.32 $\pm$ 1.49 & 94.19 $\pm$ 3.76 & 96.89 $\pm$ 1.66 & 97.19 $\pm$ 1.25 & 96.80 $\pm$ 0.91 & 50.30  & 98.30  & 83.38 $\pm$ 17.64 & 98.27 $\pm$ 0.50 & \textbf{99.23} $\pm$ 0.56 \\
        macaroni2 & 90.94 $\pm$ 1.98 & 80.47 $\pm$ 6.45 & 89.86 $\pm$ 1.44 & 94.19 $\pm$ 2.04 & 90.25 $\pm$ 2.21 & 44.70  & \textbf{97.60}  & 86.05 $\pm$ 16.68 & 87.28 $\pm$ 5.32 & 92.95 $\pm$ 5.85 \\
\bottomrule
    \end{tabular}
    }
    }
\end{table*}

\begin{table*}[t]
\caption{Anomaly segmentation on VisA-step-K4 with pixel-level AUROC. We report the mean over 5 random seeds for each measurement. The best performance is indicated with bold and the second-best with underline.}
\label{table:supp_pix_visa_k4}
    \centering
    \renewcommand{\arraystretch}{1.2}
    {\fontsize{8}{10}\selectfont
    \resizebox{\linewidth}{!}{
    \begin{tabular}{@{}l@{} | @{\hspace{2pt}}c@{\hspace{2pt}} @{\hspace{2pt}}c@{\hspace{2pt}} @{\hspace{2pt}}c@{\hspace{2pt}} @{\hspace{2pt}}c@{\hspace{2pt}} @{\hspace{2pt}}c@{\hspace{2pt}} @{\hspace{2pt}}c@{\hspace{2pt}} @{\hspace{2pt}}c@{\hspace{2pt}} @{\hspace{2pt}}c@{\hspace{2pt}} @{\hspace{2pt}}c@{\hspace{2pt}} @{\hspace{2pt}}c@{\hspace{2pt}}@{}}
    \toprule
         & PaDiM & SimpleNet & EfficientAD & HVQ & UniAD & WinCLIP & AnomalyCLIP & PatchCore & SoftPatch & \makecell{\ours\\(ours)} \\ \midrule
        candle & 92.17 $\pm$ 6.03 & 85.13 $\pm$ 9.19 & 87.24 $\pm$ 7.67 & 97.78 $\pm$ 1.33 & 95.69 $\pm$ 2.65 & 87.00  &\textbf{98.80}  & 82.17 $\pm$ 22.34 & 86.22 $\pm$ 15.30 & 98.56 $\pm$ 0.69 \\
        capsules & 73.37 $\pm$ 0.66 & 83.64 $\pm$ 4.54 & 88.48 $\pm$ 2.57 & 88.94 $\pm$ 1.14 & 82.04 $\pm$ 0.41 & 80.00  & 95.00  & 94.47 $\pm$ 0.55 & 67.11 $\pm$ 3.69 & \textbf{96.88} $\pm$ 0.39 \\
        cashew & 93.70 $\pm$ 4.97 & 92.02 $\pm$ 2.19 & 96.45 $\pm$ 0.76 & \textbf{98.07} $\pm$ 0.52 & 96.51 $\pm$ 3.14 & 84.80  & 93.80  & 78.26 $\pm$ 20.71 & 94.34 $\pm$ 4.59 & 97.74 $\pm$ 0.41 \\
        chewinggum & 93.51 $\pm$ 4.69 & 81.63 $\pm$ 5.60 & 95.61 $\pm$ 3.98 & \textbf{98.06} $\pm$ 0.36 & 97.88 $\pm$ 0.81 & 95.40  & 99.30  & 90.31 $\pm$ 10.28 & 97.50 $\pm$ 0.69 & 99.24 $\pm$ 0.20 \\
        fryum & 85.41 $\pm$ 4.52 & 79.59 $\pm$ 5.77 & 92.88 $\pm$ 3.67 & \textbf{95.65} $\pm$ 0.93 & 92.90 $\pm$ 2.43 & 87.70  & 94.60  & 86.15 $\pm$ 8.46 & 82.57 $\pm$ 3.28 & 89.69 $\pm$ 1.19 \\
        macaroni1 & 95.43 $\pm$ 4.08 & 88.33 $\pm$ 8.63 & 96.72 $\pm$ 2.59 & 97.34 $\pm$ 1.56 & 95.86 $\pm$ 2.37 & 50.30  & 98.30  & 73.56 $\pm$ 21.73 & 96.40 $\pm$ 3.07 & \textbf{99.26} $\pm$ 0.53 \\
        macaroni2 & 90.05 $\pm$ 2.20 & 74.04 $\pm$ 9.95 & 91.71 $\pm$ 3.82 & 93.46 $\pm$ 2.06 & 89.73 $\pm$ 2.26 & 44.70  & \textbf{97.60}  & 84.67 $\pm$ 15.24 & 85.76 $\pm$ 1.78 & 94.27 $\pm$ 4.41 \\
\bottomrule
    \end{tabular}
    }
    }
\end{table*}

\begin{table*}[t]
\caption{Anomaly segmentation on VisA-step-K1 with pixel-level AUROC. We report the mean over 5 random seeds for each measurement. The best performance is indicated with bold and the second-best with underline.}
\label{table:supp_pix_visa_k1}
    \centering
    \renewcommand{\arraystretch}{1.2}
    {\fontsize{8}{10}\selectfont
    \resizebox{\linewidth}{!}{
    \begin{tabular}{@{}l@{} | @{\hspace{2pt}}c@{\hspace{2pt}} @{\hspace{2pt}}c@{\hspace{2pt}} @{\hspace{2pt}}c@{\hspace{2pt}} @{\hspace{2pt}}c@{\hspace{2pt}} @{\hspace{2pt}}c@{\hspace{2pt}} @{\hspace{2pt}}c@{\hspace{2pt}} @{\hspace{2pt}}c@{\hspace{2pt}} @{\hspace{2pt}}c@{\hspace{2pt}} @{\hspace{2pt}}c@{\hspace{2pt}} @{\hspace{2pt}}c@{\hspace{2pt}}@{}}
    \toprule
         & PaDiM & SimpleNet & EfficientAD & HVQ & UniAD & WinCLIP & AnomalyCLIP & PatchCore & SoftPatch & \makecell{\ours\\(ours)} \\ \midrule
        candle & 84.28 $\pm$ 8.15 & 80.29 $\pm$ 8.82 & 76.45 $\pm$ 11.25 & 94.56 $\pm$ 3.44 & 88.46 $\pm$ 6.15 & 87.00  & \textbf{98.80}  & 88.20 $\pm$ 21.36 & 76.50 $\pm$ 12.90 & 98.18 $\pm$ 0.71 \\
        capsules & 79.63 $\pm$ 10.69 & 75.87 $\pm$ 16.78 & 87.88 $\pm$ 10.46 & 88.76 $\pm$ 9.25 & 83.30 $\pm$ 2.13 & 80.00  & 95.00  & 84.81 $\pm$ 10.74 & 75.43 $\pm$ 15.75 & \textbf{95.28} $\pm$ 0.36 \\
        cashew & 90.16 $\pm$ 6.83 & 85.16 $\pm$ 18.54 & 92.51 $\pm$ 5.09 & 93.86 $\pm$ 5.48 & 93.14 $\pm$ 5.38 & 84.80  & 93.80  & 96.75 $\pm$ 1.03 & 91.05 $\pm$ 6.64 & \textbf{97.43} $\pm$ 1.13 \\
        chewinggum & 88.79 $\pm$ 2.65 & 68.19 $\pm$ 19.17 & 83.31 $\pm$ 7.37 & 95.56 $\pm$ 1.64 & 94.26 $\pm$ 1.22 & 95.40  & 99.30  & 97.73 $\pm$ 0.12 & 94.14 $\pm$ 2.73 & \textbf{99.31} $\pm$ 0.14 \\
        fryum & 85.49 $\pm$ 7.54 & 75.45 $\pm$ 17.82 & 92.83 $\pm$ 0.69 & \textbf{93.47} $\pm$ 3.85 & 92.28 $\pm$ 4.44 & 87.70  & 94.60  & 83.20 $\pm$ 6.50 & 84.26 $\pm$ 6.65 & 89.64 $\pm$ 2.28 \\
        macaroni1 & 94.04 $\pm$ 5.81 & 78.33 $\pm$ 18.65 & 94.70 $\pm$ 5.55 & 94.64 $\pm$ 3.66 & 95.82 $\pm$ 2.37 & 50.30  & 98.30  & 71.40 $\pm$ 21.31 & 90.90 $\pm$ 10.48 & \textbf{98.64} $\pm$ 1.33 \\
        macaroni2 & 88.45 $\pm$ 0.73 & 67.73 $\pm$ 5.73 & 86.12 $\pm$ 4.12 & 90.06 $\pm$ 0.96 & 88.14 $\pm$ 2.33 & 44.70  & \textbf{97.60}  & 87.97 $\pm$ 0.86 & 82.66 $\pm$ 1.09 & 93.94 $\pm$ 1.64 \\
\bottomrule
    \end{tabular}
    }
    }
\end{table*}


\end{document}